\documentclass[a4paper,fleqn]{cas-sc}

\usepackage[numbers]{natbib}
\usepackage{amssymb}
\usepackage[utf8]{inputenc}
\usepackage{url}
\usepackage{amsfonts}
\usepackage{appendix}
\usepackage{hyperref}
\usepackage{float}
\usepackage{amsmath}
\usepackage{graphicx}
\usepackage{lipsum} 
\usepackage{subcaption}
\usepackage{algorithm}
\usepackage{algpseudocode}
\usepackage{booktabs, multirow}
\usepackage{caption}
\usepackage{soul}
\usepackage{balance}

\usepackage{acro}
\DeclareAcronym{rnn}{
	short = RNN,
	long = {Recurrent Neural Network}
}

\DeclareAcronym{dl}{
	short = DL,
	long = {deep learning}
}

\DeclareAcronym{ml}{
	short = ML,
	long = {machine learning}
}

\DeclareAcronym{pdf}{
	short = pdf,
	long = {probability density function}
}

\DeclareAcronym{erm}{
	short = ERM,
	long = {empirical risk minimization}
}

\DeclareAcronym{kmm}{
	short = KMM,
	long = {Kernel Mean Matching}
}

\DeclareAcronym{kliep}{
	short = KLIEP,
	long = {Kullback-Leibler Importance Estimation Procedure}
}

\DeclareAcronym{lsif}{
	short = LSIF,
	long = {least squares importance fitting}
}

\DeclareAcronym{atc}{
	short = ATC,
	long = {Average Thresholded Confidence}
}

\DeclareAcronym{doc}{
	short = DoC,
	long = {difference of confidences}
}

\DeclareAcronym{autoeval}{
	short = AutoEval,
	long = {Automatic model Evaluation}
}

\DeclareAcronym{nlp}{
	short = NLP,
	long = {natural language processing}
}

\DeclareAcronym{squad}{
	short = SQuAD,
	long = {Stanford Question Answering Dataset}
}

\DeclareAcronym{svm}{
	short = SVM,
	long = { Support Vector Machines}
}

\DeclareAcronym{gnb}{
	short = GNB,
	long = {Gaussian Naïve Bayes}
}

\DeclareAcronym{rf}{
	short = RF,
	long = {Random Forests}
}

\DeclareAcronym{lr}{
	short = LR,
	long = {Logistic Regression}
}

\DeclareAcronym{knn}{
	short = KNN,
	long = {K-Nearest-Neighbours}
}

\DeclareAcronym{mcc}{
	short = MCC,
	long = {Matthews coefficient of correlation}
}

\DeclareAcronym{arima}{
	short = ARIMA,
	long = {AutoRegressive Integrated Moving Average}
}

\DeclareAcronym{ems}{
	short = EMS,
	long = {energy management system}
}

\DeclareAcronym{lstm}{
	short = LSTM,
	long = {Long Short-Term Memory}
}

\DeclareAcronym{jsd}{
	short = JSD,
	long = {Jensen-Shannon divergence}
}

\DeclareAcronym{kld}{
	short = KLD,
	long = {Kullback–Leibler Divergence}
}

\DeclareAcronym{kde}{
	short = KDE,
	long = {Kernel Density Estimation}
}

\DeclareAcronym{amise}{
	short = AMISE,
	long = {asymptotic mean integrated square error}
}

\DeclareAcronym{hpo}{
	short = HPO,
	long = {hyperparameter optimization}
}

\DeclareAcronym{bo}{
	short = BO,
	long = {Bayesian optimization}
}

\DeclareAcronym{adwin}{
	short = ADWIN,
	long = {ADaptive WINdowing}
}

\DeclareAcronym{dnn}{
	short = DNN,
	long = {Deep Neural Network}
}

\DeclareAcronym{mape}{
	short = MAPE,
	long = {Mean Absolute Percentage Error}
}

\DeclareAcronym{rmse}{
	short = RMSE,
	long = {Root Mean Square Error}
}

\DeclareAcronym{cpu}{
	short = CPU,
	long = {Central Processing Unit}
}

\DeclareAcronym{gpu}{
	short = GPU,
	long = {Graphics Processing Unit}
}

\DeclareAcronym{aws}{
	short = AWS,
	long = {Amazon Web Services}
}

\DeclareAcronym{std}{
	short = STD,
	long = {standard deviation}
}

\ExplSyntaxOn
\cs_gset:Npn \__first_footerline: {
  \group_begin:
  \small \sffamily
  \__short_authors:
  \group_end:
}
\ExplSyntaxOff

\begin{document}

\shorttitle{DA-LSTM for Load Forecasting}

\shortauthors{F Bayram et~al.}

\title [mode = title]{DA-LSTM: A Dynamic Drift-Adaptive Learning Framework for Interval Load Forecasting with LSTM Networks}                      




\author[1]{Firas Bayram}[orcid=0000-0003-0683-2783]

\ead{firas.bayram@kau.se}
\author[1]{Phil Aupke}[orcid=0000-0001-9403-6175]
\ead{phil.aupke@kau.se}





\author[1,5]{Bestoun S. Ahmed}[orcid=0000-0001-9051-7609]
\ead{bestoun@kau.se}



\author[1,4]
{Andreas Kassler}[orcid=0000-0002-9446-8143]
\ead{andreas.kassler@kau.se}

\author[2]
{Andreas Theocharis}
\ead{andreas.theocharis@kau.se}

\author[3]
{Jonas Forsman}
\ead{jonas.forsman@cgi.com}

\address[1]{Department of Mathematics and Computer Science, Karlstad University, Karlstad, 65188, Sweden}
\address[2]{Department of Engineering and Physics, Karlstad University, Karlstad University, Karlstad, 65188, Sweden}
\address[3]{Advanced Analytics Solution, CGI, Karlstad, 65224, Sweden}
\address[4]{Faculty of Computer Science, Deggendorf Institute of Technology, Deggendorf, 94469, Germany}
\address[5]{Department of Computer Science, Faculty of Electrical Engineering, Czech Technical University in Prague, Prague, 16627, Czech Republic}

\begin{abstract}
Load forecasting is a crucial topic in energy management systems (EMS) due to its vital role in optimizing energy scheduling and enabling more flexible and intelligent power grid systems. 
As a result, these systems allow power utility companies to respond promptly to demands in the electricity market.
Deep learning (DL) models have been commonly employed in load forecasting problems supported by adaptation mechanisms to cope with the changing pattern of consumption by customers, known as concept drift.
A drift magnitude threshold should be defined to design change detection methods to identify drifts. While the drift magnitude in load forecasting problems can vary significantly over time, existing literature often assumes a fixed drift magnitude threshold, which should be dynamically adjusted rather than fixed during system evolution. To address this gap, in this paper, we propose a dynamic drift-adaptive Long Short-Term Memory (DA-LSTM) framework that can improve the performance of load forecasting models without requiring a drift threshold setting. We integrate several strategies into the framework based on active and passive adaptation approaches. To evaluate DA-LSTM in real-life settings, we thoroughly analyze the proposed framework and deploy it in a real-world problem through a cloud-based environment. Efficiency is evaluated in terms of the prediction performance of each approach and computational cost. The experiments show performance improvements on multiple evaluation metrics achieved by our framework compared to baseline methods from the literature. Finally, we present a trade-off analysis between prediction performance and computational costs.


\end{abstract}




\maketitle


\section{Introduction}

Global warming and the shortage of energy resources have made \ac{ems} a hot topic in the energy sector \cite{lee2016energy}. Industries have realized that developing improved \ac{ems} can potentially improve energy monitoring and budgeting. Efficient energy planning is important to reduce energy costs and optimize energy usage \cite{martin2014managing}. Enabling technologies for \ac{ems} are smart meters that measure energy consumption, which facilitates energy management at the household or building level \cite{czetany2021development}. Data collected from smart meters are valuable assets for data-driven analytics and decision-making with essential use cases, including gaining insight into consumption trends, forecasting the energy consumption load, or optimizing energy exchanges in smart microgrids, which are all core functional building blocks for \ac{ems} \cite{yildiz2017recent}.


Different prediction horizons are prominent in the load forecasting literature: short-, medium-, and long-term forecasts \cite{shah2020modeling}. Short-term forecasts refer to a prediction horizon from a few minutes to days ahead. These forecasts are essential for decision-making that involves prosumers in smart energy grids \cite{kyriakides2007short}. Medium-term forecasts treat a horizon window of weeks to a few months, which is important for scheduling power systems \cite{ringwood2001forecasting}. Lastly, long-term forecasting refers to monthly or yearly predictions, which are utilized for the maintenance planning of the grid \cite{mcsharry2005probabilistic}. 

At the top level, load forecasting aims to predict the future demand for electricity load by end-use customers. Several methods can be used for load forecasting. Recently, \ac{ml} methods have increased in popularity \cite{baliyan2015review} due to their simplicity of use. Having precise estimates of future energy demands would enable better decision-making \cite{cardenas2012load} and more effective energy planning and scheduling strategies. However, load forecasting comes with challenges of its own. One major challenge is the changes in the patterns of energy consumption by consumers over time. These changes in patterns would cause a \textit{concept drift} problem, which is induced by variation in the underlying statistical properties of the target variable \cite{lu2018learning}. This problem would lead to outdated \ac{ml} models in the predictive system after the occurrence of the change \cite{wang2019review}.

There are many reasons to change the energy consumption behaviors of customers. For example, high or changing electricity prices have greatly impacted customer behavior \cite{muratori2015residential}. Specifically, higher prices would cause what is known as \textit{demand response}, which signifies that customers change their behavior in response to high prices \cite{albadi2007demand}. Temporal or calendar factors such as year or type of day (e.g., weekday or weekend) are among the other sources of change in consumption behaviors, in addition to changes in the number of household members or the appliances used in the household \cite{khatoon2014effects}. Due to the dynamicity of the energy load consumption, the conventional \ac{ml} paradigms do not perform well and suffer from performance degradation \cite{bayram2022concept}. 

Considering the poor performance exhibited by classical \ac{ml} solutions, deep learning-based approaches have been widely adopted in recent research on interval load forecasting since they achieve superior performance \cite{dong2021electrical}. 
Specifically, \ac{lstm} is an effective algorithm that demonstrated high performance in the load forecasting problem \cite{siami2018comparison}. Moreover, to maximize the efficiency of the \ac{dl} models, an adaptive mechanism is fostered to cope with the changes in load consumption. This adaptation mechanism is set to automatically update the model and adjust it to the new energy usage pattern that presents concept drift \cite{hou2021novel}. In practice, concept drift is usually tracked actively or passively by the probability distribution that generates the data stream \cite{webb2016characterizing}. However, the main challenge in drift detection is determining a change magnitude threshold that exerts a great influence on overall predictive performance \cite{wares2019data}. The change threshold should not be fixed, but should be tuned according to the present conditions of the system \cite{liu2022concept}.

In this paper, we build an interval load forecasting learning framework based on dynamic drift adaptation for \ac{lstm} networks, namely DA-LSTM. We design different load forecasting solutions based on passive and active drift adaptation techniques. We highlight the advantages and disadvantages of each solution.
Overall, the primary contributions and findings of this paper are summarized as follows:

\begin{enumerate}
        \item A novel drift-adaptive LSTM (DA-LSTM) learning framework is proposed for interval load forecasting, which can be integrated with passive and active drift adaptation techniques.
        \item A dynamic active drift detection methodology that identifies the change point in a consumer's behavior without fixing a drift magnitude threshold.
        \item An adaptive \ac{lstm} network is designed to handle concept drift by quickly adapting to the new trend in load consumption while retaining the learned consumption patterns.
        \item An extensive evaluation is conducted against baseline models from the literature to demonstrate the effectiveness of the proposed DA-LSTM framework.
        \item A trade-off analysis between the different adaptation strategies is performed based on prediction performance and computational cost to suggest the adoption of the appropriate approach.
\end{enumerate}


The remainder of the paper is organized as follows. In Section \ref{sec:related}, we review the related work on load forecasting and Section \ref{sec:background} introduces the background of the problem. Section \ref{sec:proposed} introduces our novel drift-adaptive LSTM approach. Section \ref{results} shows the experimental evaluation of our approach and demonstrates results, and Section \ref{sec:conclusion} concludes the paper.


\section{Related work}
\label{sec:related}

\subsection{DL for Residential Load forecasting}

The building characteristics and socio-economic variables are the most commonly used exogenous variables for building and occupancy prediction. Artificial neural network (ANN), bottom-up, time series analysis, regression, and \ac{svm} are the most often used load forecasting models, according to a survey by Kuster \textit{et al.} \cite{survey2017}. The survey has also concluded that most regression models are used for long-term prediction, one year or more, while ML-based algorithms, such as ANN, especially \ac{dl}, and \ac{arima} \cite{arima2020}, are commonly used for short-term prediction \cite{hernandez2014artificial}, which is also what we aim for. Alternative methods in the literature consider additional variables to improve forecasting accuracy. For example, the model provided by Hong \textit{et al.} \cite{hong2020deep} leverages iterative ResBlocks in a \ac{dnn} to learn the spatial-temporal correlation among different types of user's electricity consumption habits.



DL algorithms such as \acp{rnn} and Convolutional Neural Networks (CNN) have demonstrated significant efficiency. Nevertheless, these approaches use offline learning: they are taught only once and miss the potential to learn from newly arriving data. Sehovac et al. \cite{sehovac2020deep} proposed Sequence to Sequence Recurrent Neural Network (S2S \ac{rnn}) with attention to load forecasting. The concept behind S2S \ac{rnn} is based on adopting an attention mechanism \cite{vaswani2017attention}, often utilized in language translation, to load forecasting to improve accuracy. The attention mechanism in the method is added to ease the connection between the encoder and decoder. A hybrid approach between \ac{rnn} and Principal Component Analysis (PCA) technique has been proposed  by Veeramsetty \textit{et al.} \cite{veeramsetty2022short} in a short-term load forecasting problem. The approach can capture the temporal resolution diversity using a heterogeneous input structure with PCA. The PCA-based summarized input features are used as input to an RNN model.

\ac{lstm} network is a particular type of \ac{rnn} that has been effectively utilized in load forecasting problems. Zang \textit{et al.} \cite{zang2021residential} have combined \ac{lstm} and attention mechanism for load forecasting of residential households. The approach constructs pools of users based on mutual information to increase the diversity of data used to train \ac{lstm} networks. To leverage the advantages of both networks, an integration between CNN and \ac{lstm} has been frequently used in the load forecasting literature \cite{eskandari2021convolutional, rafi2021short, goh2021multi, somu2021deep}. CNN exhibits high capabilities in feature learning, while \ac{lstm} can handle short- and long-term temporal dependencies between time steps. Additionally, recent studies have demonstrated the effectiveness of employing advanced neural networks for load forecasting such as temporal convolutional network (TCN) \cite{tang2022short, song2020hourly}, and Restricted Boltzmann Machine (RBM) \cite{xu2022new}. However, changes in customer behaviors can lead to deterioration in the performance of the learning algorithm. Therefore, load forecasting approaches have been integrated with a drift-adaptive methodology to cope with the change \cite{azeem2022deterioration}.


\subsection{Load Forecasting with Concept Drift}

Only a few approaches have accommodated concept drift adaptation techniques in the load forecasting models. In recent studies, Fekri \textit{et al.} \cite{onlineRNN2021} proposed an online adaptive \ac{rnn} technique for load forecasting that considers concept drift. The model can learn and adapt to changing patterns as they emerge. This is done by adjusting the \ac{rnn} weights online based on fresh data to retain time dependencies. The on-the-fly adjustment of the \ac{rnn} parameters is activated in the event of performance degradation. Similarly, Jagait \textit{et al.} \cite{Jagait2021Load} has proposed an adaptive online ensemble with \ac{rnn} and \ac{arima} for load forecasting in the presence of concept drift. The adaptation to changes is made by adding Rolling \ac{arima} to the ensemble. Another approach that uses incremental ensemble learning has been presented in \cite{Grmanova2016Incremental}. The model uses a heterogeneous learning process to build an ensemble that deals with seasonality and concept drift. Fenza \textit{et al.} \cite{fenza2019drift} presented a drift-aware solution to distinguish the anomaly behavior of customers from the regular pattern. The change is detected based on the standard deviation of the prediction error in the last week.

In a different approach proposed by Ji \textit{et al.} \cite{Ji2021Enhancing}, the \ac{adwin} algorithm \cite{bifet2007adwin} has been utilized in short-term load forecasting problems to detect concept drift in a model updating method. Load forecasting methods that rely on calendar or weather information trained on historical data fail to capture significant break induced by lockdown and have performed poorly since the COVID-19 pandemic began. The article in \cite{pandemicDrift2021} anticipates the electricity demand in France during the lockdown period, demonstrating its ability to significantly minimize prediction errors compared to conventional models. The method uses Kalman filters and generalized additive models to produce an accurate and rapid forecasting strategy to respond to the sudden shift in data. However, existing approaches that rely on drift adaptation require setting a drift threshold to trigger the alarm, which is a drawback, as it primarily influences predictive performance \cite{wares2019data}.  

With the limited number of studies that exploit drift handling techniques in the load forecasting area, we propose a novel framework that combines drift adaptation approaches with \ac{lstm} networks. We incorporate a dynamic drift detection technique that does not require a pre-defined drift threshold into the framework. Instead, the method checks how extreme the drift magnitude is by leveraging the distribution of the drift magnitudes. In this way, there is no need to define a drift threshold that can be difficult to determine or fix throughout the learning process. Furthermore, the proposed framework can respond rapidly to drifts by taking advantage of the most recent patterns. At the same time, the framework can retain the learned knowledge, which may re-appear in the future. We performed extensive experimental analysis on a real-world energy consumption dataset. The result demonstrates the strength of our framework in prediction accuracy compared to traditional techniques.

\section{Background}
\label{sec:background}

The statistical properties of the data streams often do not remain stable over time and changes are likely to occur, a phenomenon known as \textit{concept drift} \cite{Tsymbal2004Problem}. Statistical tests are usually used to monitor and detect concept drift. Concept drift typically leads to a reduction in the accuracy of forecasting models. Adaptation strategies have typically been implemented to cope with performance loss that update the models to cope with drift \cite{ditzler2015learning}. Concept drift adaptation strategies can be categorized into \textit{passive}, also known as \textit{blind}, and \textit{active}, also known as \textit{informed}, methods \cite{song2021segment}. A passive adaptation denotes the drift adaptation that does not include drift detection techniques, and the predictor is regularly updated \cite{khamassi2018discussion}. An active adaptation is the drift adaptation strategy that is actively triggered once a concept drift alarm is signaled from a concept drift detector \cite{dong2021drift}.

In the load forecasting context, changes in consumption behavior may have multiple reasons, including changes in the number of household members, weather variations, or adding new electrical appliances and devices such as gaming PCs or electric vehicles. Thus, the concept drift solution strategy should be mainstreamed in load forecasting problems.  This section summarizes the background on the change-point detection problem and the adaptive mechanism introduced in the \ac{lstm} network to cope with the drift. 

\subsection{Change-Point Detection}
\label{sec:change-point}

To identify variations in load consumption behavior, change-point detection methods are employed to test whether load consumption data have followed a change at a specific time point $t$. The following subsections present the definitions and notations for the change-point detection problem. We then illustrate the methods used to detect changes in load consumption patterns using distributional similarity measures.

\subsubsection{Problem Formulation}
\label{sec:statement}
\begin{figure*}
\centering
\includegraphics[width=1\textwidth]{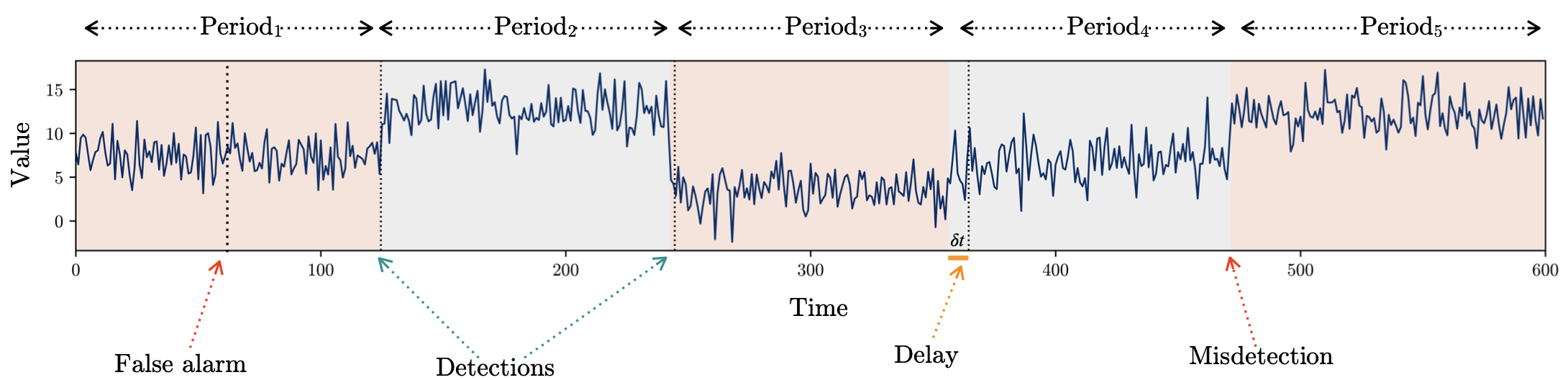}
\caption{Change-point detection in time-series data}
\label{fig:cpd_plot}
\end{figure*}

\textit{Change-point detection}, also known as \textit{drift detection}, refers to the techniques used to identify change points of time-series data (such as electrical load) where a significant change has occurred in the underlying probability distribution that generates the data points \cite{liu2013change}. The change detectors are usually coupled with a predictive system. When the change exceeds the significance level, the detectors signal an alarm and evoke the learner to be updated or replaced in the forecasting system \cite{krawczyk2017ensemble}. Practically, the change signal is delayed by at least one time point \cite{song2021segment}. The mechanism of change-point detection algorithms is illustrated in Fig. \ref{fig:cpd_plot}.  Different performance indicators for the change-point detection algorithms are annotated in the figure. For example, the detection delay $\delta t$ is the time between drift occurrence and detection. Another indicator is the misdetection rate, which quantifies the number of changes missed by the algorithm. In contrast, the false alarm rate is the ratio of incorrectly detected change points compared to real drift occurrences. However, in real-world datasets, the ground-truth information of changes is usually indeterminate \cite{haug2021learning}. Therefore, the performance of change-point detection methods is usually assessed by the predictive performance of the ML model along with the associated adaptation costs.

In load forecasting problems, data are typically recorded as time-series vectors, and it is fundamental to divide these data vectors into time-series samples for the purpose of drift detection. Analogously to the formulation presented by Liu \textit{et al.} \cite{liu2013change}, we define the following notation of our problem. Let $\mathbf{Y}(t) \in \mathbb{R}^{k}$ be a sequence of univariate time series of load consumption observations with length $k$ at time $t$:\\
$\mathbf{Y}(t):=\left[{y}(t-k+1),\ldots,{y}(t-1),{y}(t)\right]^{\top} \in \mathbb{R}^{k}$,\\
where $y(t) \in \mathbb{R}$ is a single load consumption observation with an order that signifies the temporal dependency between the observations of each time-series sequence, $^{\top}$ denotes the transpose operation of the vector. By convention, since the granularity of the change-point detection analysis of time-series data is the entire sequence of load consumption observations $\mathbf{Y}(t)$ rather than a single observation $y(t)$, the term \textit{“sample”} is adopted to refer to the overall sequence of observations $\mathbf{Y}(t)$, instead of the single observation $y(t)$ \cite{kawahara2009change, liu2013change}. Therefore, the set of $n$ consecutive load consumption samples at time $t$ is defined as:\\
$\mathbb{Y}(t):=\{\mathbf{Y}(t-n+1), \ldots, \mathbf{Y}(t-1),  \mathbf{Y}(t)\}$. The task of change-point detection is to identify the (dis)similarity between two samples at time $t$. Typically, the (dis)similarity score $D(t)$ is calculated using the probability distributions of the two samples. This similarity score will be used as an indicator to diagnose the status of the time-series data.  In load forecasting literature, existing methods that employ drift detectors compares $D(t)$ to a pre-defined threshold $\lambda$ to detect change points \cite{zhao2020review, li2022aws}. However, energy consumption patterns are characterized by high volatility \cite{kong2017short}. Therefore, the threshold $\lambda$ must be dynamic to cope with the changing nature of the energy consumption patterns.


\subsubsection{Divergence-Based Similarity Score}
\label{sec:divergence}

As defined in Section \ref{sec:statement}, change-point detection methods rely on identifying the significance of the change incurred in the data-generating probability distributions. A prevalent approach to measure the similarity between two or more probability distributions of load consumption data is to calculate the divergence between these distributions \cite{lee2000measures}. Higher magnitudes of the divergence values indicate a more significant dissimilarity between probability distributions.

In our settings, we are interested in measuring the similarity at time \textit{t} between the probability distribution of two consecutive samples of load consumption data to inspect the drift at \textit{t}. The similarity score $D(t)$ will be calculated based on the comparison between a reference sample of load consumption $\mathbf{Y}_{rf}= \mathbf{Y}(t-1)$ and the most recent sample of load consumption in which the similarity will be tested $\mathbf{Y}_{ts}= \mathbf{Y}(t)$. Note here that the samples are set to be segmented according to a pre-defined window size. Windowing techniques generally require defining the window size based on the number of data points or the time interval \cite{Gama2014Survey}. The selection of window size will be explained in Section \ref{sec:granularity}. 

\ac{jsd} is one of the most popular metrics used to measure the magnitude of the distributional change that the time-series data have followed at a specific time point $t$ \cite{endres2003new}. \ac{jsd} is a well-grounded symmetrization of the well-known \ac{kld} metric.  Liu \textit{et al.} \cite{liu2013change} showed that using a symmetric divergence metric leads to greater change-point detection performance. The distance between two probability distributions using \ac{jsd} is defined as \cite{lin1991divergence}:
\begin{equation}
\mathrm{JSD}(P_{rf}\| P_{ts}) = \mathrm{JSD}(P_{ts}\| P_{rf}) := \frac{1}{2}\left(\mathrm{KL}\left(p_{rf}\|\frac{p_{rf}+p_{ts}}{2}\right)+ \mathrm{KL}\left(p_{ts}\| \frac{p_{rf}+p_{ts}}{2}\right)\right),
\end{equation}
where $P_{rf}$ and $P_{ts}$ are the probability distributions for the reference and test samples $\mathbf{Y}_{rf}$ and $\mathbf{Y}_{ts}$, respectively, the function $\mathrm{KL}$ represents \ac{kld} and is given by:
\begin{equation}
\mathrm{KL}\left(P \| Q\right):=\int p(\mathbf{Y}) \log \left(\frac{p(\mathbf{Y})}{q(\mathbf{Y})}\right) d \mathbf{Y}.
\end{equation}
Therefore, \ac{jsd} can be written as:
\begin{equation}
\label{entropy}
\mathrm{JSD}(P_{rf}\| P_{ts}) = \mathrm{JSD}(P_{ts}\| P_{rf}) := H\left(\frac{p_{rf}+p_{ts}}{2}\right)-\frac{H(p_{rf})+H(p_{ts})}{2},
\end{equation}
where the function $H$ denotes Shannon's entropy:
\begin{equation}
H(p)=-\int p(\mathbf{Y}) \log p(\mathbf{Y}) d \mathbf{Y}.
\end{equation}


In addition to its symmetry, \ac{jsd} exhibits several pertinent properties: It is always defined and bounded in the interval $[0,1]$ for two probability distributions, while the value of \ac{kld} can diverge to infinity and can take values in the interval $[0, \infty)$,  and the square root of \ac{jsd} satisfies the triangle inequality \cite{endres2003new}. For the abovementioned properties, we use the square root of \ac{jsd} as the distance metric in the change-point detection experiments.


\subsubsection{Nonparametric Density Estimation}
\label{sec:density}
Density estimation is the technique used to recover the probability density function that generates the dataset.
As we can see from Eq. \ref{entropy}, to calculate the \ac{jsd} value we first need to find the probability distribution $p(\mathbf{Y})$ to determine the distance between the samples. Several methods to estimate the probability distribution can be found in the literature. Histograms and \ac{kde} methods are the most widely used nonparametric approaches to reconstruct the underlying probability density function using a given dataset. 

Although, in general, \ac{kde} has a slower convergence rate of magnitude $N^{-1}$ than traditional histograms \cite{burke2017kernel}, where $N$ is the number of data points, histograms are susceptible to exhibit a lower convergence rate in dynamic environments that are characterized by changing variance. When the dispersion of the data changes, the number of bins should be varied as suggested by \cite{banerjee2012kernel, burke2017kernel}. However, \ac{kde} suffers from the so-called phenomenon \textit{curse of dimensionality}, which makes the convergence rate very slow when the dimension of the problem is large \cite{chen2017tutorial}. The optimal convergence rate of \ac{kde} is $O\left(N^{-\frac{2}{d+4}}\right)$, where $d$ is the dimension of the data. We have adopted the \ac{kde} method in estimating the density because this phenomenon will not affect the performance of our proposed solution since the time-series samples of our dataset are \textit{univariate}.

To produce the density profile, \ac{kde} places a kernel at each data point $y_i$, and then sums these individual kernels together to obtain the final density estimate. This method will render a smooth curve, which is one of the salient advantages of the \ac{kde} method \cite{bouezmarni2005consistency}. At regions with a high density of data points, the \ac{kde} will yield a large value, because many points will contribute to the sum value. However, it will yield a low value for regions with only a few data points.   With $\left(y_{1}, y_{2}, \ldots, y_{n}\right)$ being a sample of $n$ observations whose underlying probability distribution is to be estimated, where $y_{i} \in R^d$. The standard \ac{kde} function is formally expressed as follows \cite{silverman2018density}:
\begin{equation}
\widehat{p}_{n}(y)=\frac{1}{n h^{d}} \sum_{i=1}^{n} K\left(\frac{y-Y_{i}}{h}\right),
\label{eq:kde}
\end{equation}
where $K$ is a smooth kernel function $K: \mathbb{R}^{d} \mapsto \mathbb{R}$, $h>0$ is the bandwidth or the smoothing parameter. As has been remarked in the literature, the choice of the kernel function is not instrumental for \ac{kde} \cite{wasserman2006all}, and the difference in the estimation error is considered negligible \cite{chen2017tutorial}.
Thus, for selecting the kernel function, we opted to use the Gaussian kernel, which is the most common kernel function, given by the form:
\begin{equation}
K\left(\frac{y-Y_{i}}{h}\right)=\frac{1}{\sqrt{2 \pi}} e^{-\frac{\left(y-Y_{i}\right)^{2}}{2 h^{2}}}.
\label{eq:gaus_ker}
\end{equation}

On the contrary, as can be noticed in Eq. \ref{eq:kde}, the density function is strongly affected by the selection of the bandwidth parameter. Therefore, different bandwidth values would give different density function values as it controls the \textit{smoothness} or \textit{roughness} of the density estimate (see Eq. \ref{eq:gaus_ker}). Many procedures attempt to adaptively determine the optimal bandwidth parameter \cite{scott2015multivariate}, mainly by minimizing the \ac{amise} estimations of \ac{kde} \cite{chiu1996comparative}. The drawback of such automated procedures for adaptive smoothing would most likely lead to the selection of different values of bandwidth parameters for different samples. This means that the amount of smoothing varies between different time intervals or households in our problem, and thus the comparison between the data distributions is incoherent. In this case, the divergence calculation would be futile because the densities are constructed with different amounts of smoothing. To overcome this drawback, we have used a fixed bandwidth value across all the samples to extract homogeneous \ac{kde} functions and make the distance calculation valid. Conventionally, bandwidth is set according to domain expertise for load forecasting and is customized for the specific problem using data-driven statistics \cite{ye2019data}. In this paper, we have set the bandwidth value to \textbf{10} based on empirical experience gained from observing the standard deviation of the load consumption across all customers.

\subsection{Conventional LSTM}
\label{sec:convlstm}

The conventional \ac{lstm} \cite{patterson2017deep} is a sort of \ac{rnn} but with additional long-term memory. The standard \acsp{rnn} uses recurrent cells such as sigma, which can be expressed as follows \cite{yu2019review}:
\begin{equation}
\begin{aligned}
&h_{t}=\sigma\left(W_{h} h_{t-1}+W_{x} x_{t}+b\right), \\
&y_{t}=h_{t+},
\end{aligned}
\end{equation}
where $x_{t}$, $h_{t}$ and $y_{t}$ represent the input, recurrent information, and the output of the cell at time $t$, respectively, $W_{h}$ and $W_{x}$ are the weights, and $b$ is the bias. The time-series sample for $x_{t}$ and $y_{t}$ are derived from $\mathbf{Y}(t)$. Although the usage of standard \acsp{rnn} cells provided success in problems such as sentiment analysis or image classification (\cite{cho2014learning},\cite{shewalkar2019performance}), they typically cannot treat long-term dependencies well.  \ac{lstm} networks can remember values from earlier stages to use in the future, which deals with the vanishing gradient problem \cite{hochreiter1997long}. This problem occurs because the network cannot backpropagate the gradient information to the input layers of the model due to activation functions. The sigmoid function, for example, normalizes large input values in a space between 0 and 1. Therefore, a large change in the input will cause a small change in the output. Therefore, the derivative becomes small and possibly vanishes \cite{sundermeyer2012lstm}. To deal with this problem, \cite{hochreiter1997long} introduced gates into the cell.


The conventional \ac{lstm} cell features three gates (input, forget, and output), a cell, block input, and an output activation function. The cell output is recurrently connected back to the cell input and all the gates. The forget gate was not part of the initial LSTM network, but was proposed by Gers \textit{et al.} \cite{gers2000learning} to allow the \ac{lstm} to reset its state. The three gates regulate the flow of information associated with the cell. 

The input gate combines the current input $X_t$, the output of the prior \ac{lstm} cell $h_{(t-1)}$ and the cell state $C_{(t-1)}$. The following equation illustrates the procedure \cite{yu2019review}:
\begin{equation}
i^{(t)}=\sigma\left(W_{i} x^{(t)}+R_{i} h^{(t-1)}+p_{i} \odot c^{(t-1)}+b_{i}\right),
\end{equation}
where $\odot$ denotes point-wise multiplication of two vectors, $W_i$, $R_i$, and $p_i$ are the weights associated with $x_t$, $h_{(t-1)}$, and $C_{(t-1)}$, respectively. $b_i$ represents the bias vector associated with this component. The prior \ac{lstm} layer determines which information should be retained in the cell states $c_t$. This includes the selection of candidate values $z_t$ that could be added to cell states and activation values $i_t$ of the input gates.

The forget gate determines which information should be removed from its previous cell states $C_{(t-1)}$. Therefore, the activation values $f_t$ are calculated based on the current input $x_t$, the outputs $h_{(t-1)}$ and the state $C_{(t-1)}$ of the memory cells at the previous time step $(t-1)$. 
\begin{equation}
f^{(t)}=\sigma\left(W_{f} x^{(t)}+R_{f} y^{(t-1)}+p_{f} \odot c^{(t-1)}+b_{f}\right),
\end{equation}
where $W_f$, $R_f$, and $p_f$ are the weights associated with $x_t$, $h_{(t-1)}$ and $C_{(t-1)}$, respectively, while $b_f$ denotes the bias vector.

The cell state combines the input values of the block $z_t$, the input gate $i_t$, and the forget gate $f_t$, with the previous cell value: 
\begin{equation}
c^{(t)}=z^{(t)} \odot i^{(t)}+c^{(t-1)} \odot f^{(t)}.
\end{equation}

The output gate combines the current input $x_t$, the output of the previous unit $h_{(t-1)}$, and the cell value $C_{(t-1)}$ in the last iteration: 
\begin{equation}
o^{(t)}=\sigma\left(W_{o} x^{(t)}+R_{o} h^{(t-1)}+p_{o} \odot c^{(t)}+b_{o}\right),
\end{equation}
where $W_o$, $R_o$ and $p_o$ are the weights associated with $x_t$, $h_{(t-1)}$ and $C_{(t-1)}$, respectively, while $b_o$ denotes the bias weight vector. 


\subsection{Adaptive LSTM} 
\label{sec:adaptive}

Inspired by incremental learning techniques, which are useful in solving drift problems \cite{zang2014comparative}, we continuously adapt the \ac{lstm} model to pertain to the previously acquired knowledge and update the model based on the most recent data that represent the current trend in load consumption. Incremental learning is an \ac{ml} paradigm in which the learning process continually evolves whenever new samples emerge. Furthermore, incremental learning adjusts what has been learned according to these newly available samples \cite{ade2013methods}. The literature provides different definitions for incremental learning (\cite{zhou2002hybrid,xu2016three,lange2003formal,giraud2000note}). In this paper, we adopt a universally accepted setting for incremental learning that satisfies the following conditions (\cite{lange2003formal,giraud2000note}):
\begin{itemize}
        \item Knowledge obtained previously should be preserved;
        \item Learning new knowledge from new data should be possible;
        \item Knowledge of previous historical data is not required when updating the model; 
        \item Changes in the characteristics of the new data should be learned.
\end{itemize}

Incremental learning algorithms can be categorized into two main approaches: the growing or pruning of the model architectures \cite{hung2019compacting} and the controlled modification of the learner weights \cite{xing2016self,singh2021enhanced}. In this paper, we adopt the latter approach, which is appropriate for load forecasting tasks. 
The main motivation for our choice is that old consumption behavior usually reoccurs in the future due to seasonality \cite{he2017load}. Therefore, modifying the current models while preserving the previously learned knowledge would be more favorable than completely forgetting the historical knowledge. The adaptation is divided into two main steps: the \ac{hpo} technique and the adaptation of the \ac{lstm} model.



\subsubsection{Hyperparameter Optimization (HPO)} \label{sec:hpo}

\ac{hpo} is a technique to find the best combination of hyperparameters that optimize the performance of \ac{ml} model \cite{feurer2019hyperparameter}. Tuning the hyperparameters for specific problems leads to increased performance \cite{melis2017state}. 




Different techniques are used in the literature to automatically find the optimal hyperparameters. Grid search and random search are the most basic \ac{hpo} methods. However,  they suffer from unawareness of past evaluations, which often leads towards sub-optimal hyperparameters \cite{montgomery2017design}. \ac{bo} \cite{bergstra2011algorithms} overcomes this by keeping track of previous evaluations by forming surrogate models to map hyperparameters to a probability $P$ of the objective function score: 
\begin{equation}
P(\text { score } \mid \text { hyperparameters }).
\end{equation}

\ac{bo} uses probabilistic surrogate models for the objective function because they are easier to optimize than the actual objective function. The process of \ac{bo} is described in Algorithm \ref{algo1}. The probabilistic model $P_{model}$ is initialized on $f(.)$ a priory. For each iteration, the best set of $x_{\star}$ is found for the current $P_{model}$ model, and the model $score$ for the set $x_{\star}$ is determined and $P_{model}$ is updated \cite{bergstra2011algorithms}. 

Since we apply incremental learning techniques in this work, using \ac{hpo} for every hyperparameter available in \ac{lstm} networks is not applicable. Structural hyperparameters, such as the number of layers and the size of the hidden layer affect the weights of the model. Changing those requires a complete retraining of the surrogate model, which invalidates the purpose of surrogate models, in general. However, non-structural hyperparameters such as learning and dropout rate do not require a complete re-training of the model, which makes their tuning applicable in this case \cite{onlineRNN2021,greff2016lstm}. Table \ref{tab:hp} illustrates the ranges of the non-structural hyperparameters used for the \ac{hpo} in the experiments.

\begin{table}[!hbtp]\centering
        \caption{Search space for hyperparameter values}\label{tab:hp}
        \setlength{\tabcolsep}{5pt}
        \scriptsize
        \begin{tabular}{lr}\toprule
            Hyperparameter& Values \\\midrule
            Learning rate                     & LR=[0.0001, 0.001, 0.01] \\
            Dropout rate                      & DR=[0, 0.1, 0.5] \\
            Number of units in the LSTM layer & $N_{U}$=(32: 32: 512) \\
            \bottomrule
        \end{tabular}
\end{table}

\begin{algorithm}[t]
\caption{Bayesian Hyperparameter Optimization}\label{alg:cap}
\begin{algorithmic}[1]
\State $P_{model} \gets Surrogate(f(x))$
\State $x_{\star} \gets []$
\While{$i < maxIterations$}
    \State $x_{\star} \gets P_{model}(score,hyperparameters)$
    \State $score \gets f(x_{\star})$
    \State $P_{model} \gets Update(P_{model},score)$
\EndWhile
\end{algorithmic}
\label{algo1}
\end{algorithm}


\subsubsection{Adaptation of LSTM Model}
\label{sec:update_lstm}

As mentioned above, we were inspired by incremental learning techniques to adapt the \ac{lstm} model. To suit the receiving data for the model, we followed the batch-wise approach for incremental \ac{lstm} used in \cite{lemos2022incremental}. Since \ac{lstm} expects the dataset as batches for the training, we divide it into multiple sub-batches for incremental processing.  

The stored weights $W_t$ of the baseline model \ac{lstm} retain the knowledge preserved from the previous historical data. After an update signal is triggered, these weights must be adapted to the new batch of data points and the corresponding updated non-structural hyperparameters. To preserve already learned knowledge and reduce training time, stored weights $W_t$ are updated only with new data points, excluding previous historical data. This is enabled by the possibility of continuing the training at the point of the storage of the weights $W_t$. The new weights $W_{(t+1)}$ of the adapted model are stored later to be used for the next update signal.


\section{Proposed Drift-Adaptive \ac{lstm} (DA-LSTM)}
\label{sec:proposed}
The \ac{lstm} algorithm was integrated into a drift-adaptive learning framework using adaptation techniques. The techniques are active and passive approaches based on the methodology implemented to trigger the adaptation of \ac{lstm} networks. This section explains the selection of drift detection granularity and the drift-adaptive LSTM (DA-LSTM) approaches.
\subsection{Drift Adaptation Granularity}
\label{sec:granularity}

As discussed in Section \ref{sec:change-point}, change-point detection requires comparing the distributions of different samples. The samples should be partitioned according to a specific windowing strategy. Therefore, the data distribution is built for these windows. The main ingredient for such windowing strategies is the window size that represents the checking points to monitor the drift occurrence.  The selection of window size significantly affects drift analysis and should be carefully decided \cite{webb2018analyzing}. In load forecasting, the load consumption trends can be monitored across spatial or temporal scales \cite{jain2014forecasting}. But since our dataset, as will be discussed in Section \ref{sec:dataset}, does not have spatial information about the households, we have used temporal granularity to divide the time-series data.

\begin{figure*}
\centering
\includegraphics[width=1\textwidth]{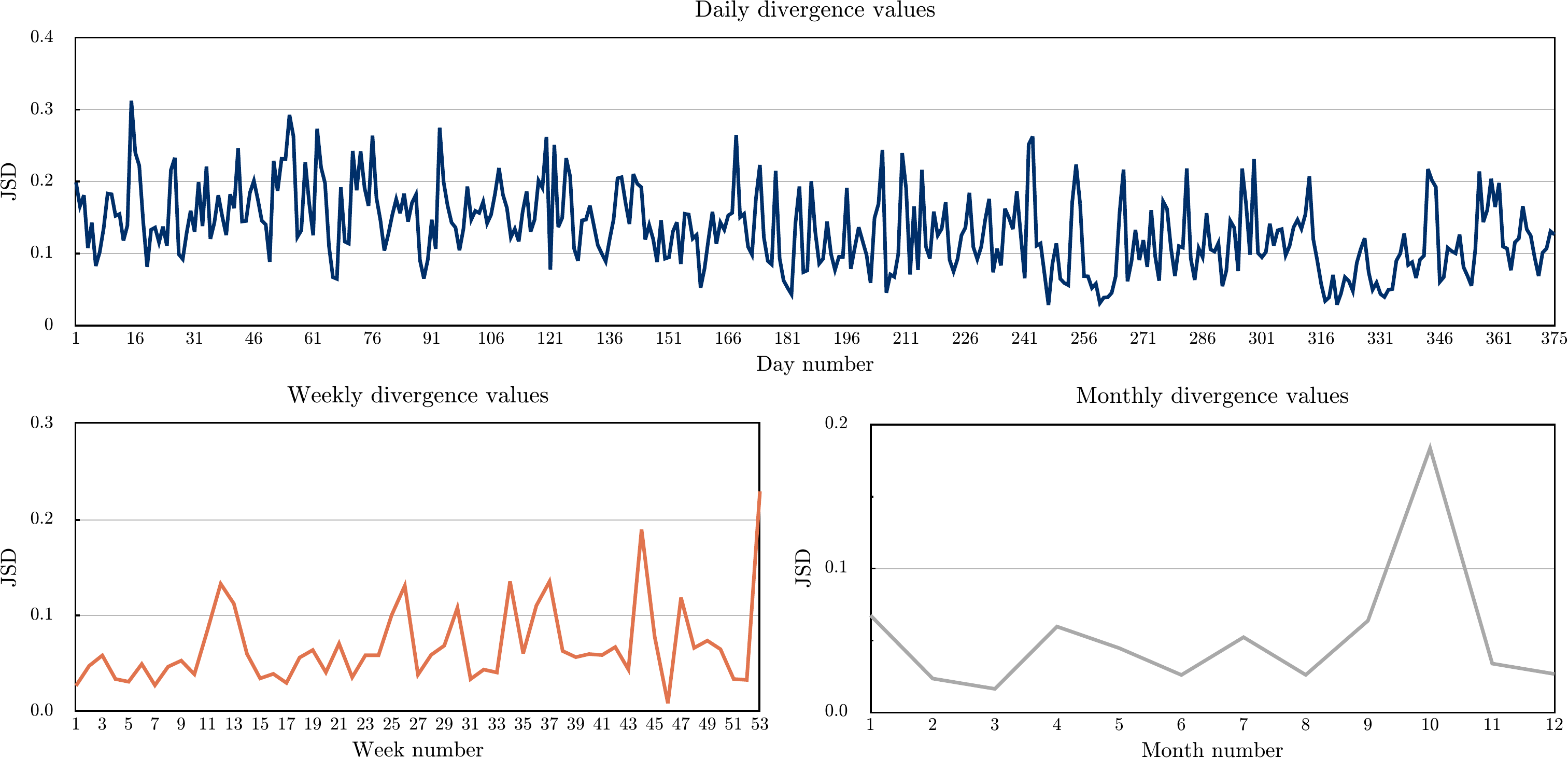}
\caption{JSD divergence magnitude using different temporal granularity levels for the households}
\label{fig:div_mag}
\end{figure*}

Different levels of granularity can be defined to check the efficacy of the adaptation methods. For good performance, it is important to determine the optimal temporal granularity for drift analysis.
Short-term forecasting was shown to deliver the best performance among prediction horizons \cite{long2014analysis}. To support this argument, we have compared the drift magnitude for the household consumption data according to different temporal granularity levels: daily, weekly, and monthly. The \ac{jsd} values of all customers are averaged, and the results are reported in Fig. \ref{fig:div_mag}. As we can see, the daily granularity shows the highest level of drift magnitude by showing the most significant divergence values. For this reason and to avoid overriding drift occurrences, we have adopted daily granularity in our adaptation solution.

\subsection{Passive Adaptation Approach}



The passive drift adaptation scheme is similar to incremental learning from a technical point of view \cite{song2021learning}. The primary analogy is that both mechanisms work without explicitly detecting a change, i.e., no drift detector component is involved, and the learning process evolves with the data arrival. Following the integration of this adaptation mechanism with the learning process, the model is updated after a specific time interval or chunk size. This periodicity is identified to sufficiently resemble the latest consumption trend and thus can handle the potential change in the time series.

Algorithm \ref{algo:passive} outlines the passive drift adaptation approach, which periodically updates the \ac{lstm} networks. The periodicity of the model update assumes a \textit{daily} granularity, as it shows the highest drift significance, as discussed in Section \ref{sec:granularity}. First, the \ac{lstm} model is initialized using the historical time-series data. A daily update signal is sent to the \ac{lstm} model to be updated according to the method explained in Section \ref{sec:adaptive}. This update mechanism is designed to incorporate the most recent consumption trends represented by the most recent data. 

However, one of the criticisms of this approach is derived from the constant update of the trained learner. This setup of constant updates might accommodate unnecessary updates when the time-series data have been stationary in some intervals. The unnecessary updates would cause an avoidable computational cost and resources \cite{ditzler2012incremental}. To overcome this drawback, we present an active approach, discussed in the next section, that only updates the model after drift detection in the time-series data. 

\begin{algorithm}[t]
\caption{Passive drift adaptation algorithm}
\begin{algorithmic}[1]
\Require Historical time-series samples $\mathbb{Y}(t-1)$
\Ensure  Learner $f^{(t)}$
\State \textbf{Initialize:} $f^{(t-1)}\big(\mathbb{Y}(t-1)\big)$
\For {$\mathbf{Y}(t)$, $t=2,\dots$}
\State Update $f^{(t)} \leftarrow f^{(t-1)}\big(\mathbf{Y}(t)\big)$
\State Predict $f^{(t)}\big(\mathbf{Y}(t+1)\big)$
\State Calculate $E_{t+1}$
\EndFor
\end{algorithmic}
\label{algo:passive}
\end{algorithm}

\subsection{Active Adaptation Approach}
Drift detectors form the foundation of active adaptation solutions. The detectors continuously monitor the time-series data and perform a set of statistical tests to identify the changes in the data. In this approach, we employ a change-point detection based on Jensen-Shannon divergence to trace the changes in the time-series data distribution, see Section \ref{sec:change-point}. Once the change exceeds a certain significance level $\tau$, the adaptation phase is activated by sending an update signal to the \ac{lstm} networks. The \ac{lstm} adaptation is carried out as specified in Section \ref{sec:adaptive}.

One of the main challenges to create a drift detection method is selecting a suitable drift threshold \cite{barros2018large}. Setting a low drift magnitude threshold would lead to an increased false alarm rate. However, setting a high threshold value would cause drift misdetections. In a recent study \cite{liu2022concept}, the authors pointed out that the selection of a suitable drift threshold could have a greater impact on performance than the detection algorithm. Moreover, the authors have demonstrated that the drift threshold should not be fixed throughout the learning process and must follow the dynamicity of the data based on the current conditions. Motivated by these observations, we have developed a dynamic mechanism to efficiently signal a time-series data distribution change. 

\begin{figure*}
\centering
\includegraphics[width=0.9\textwidth]{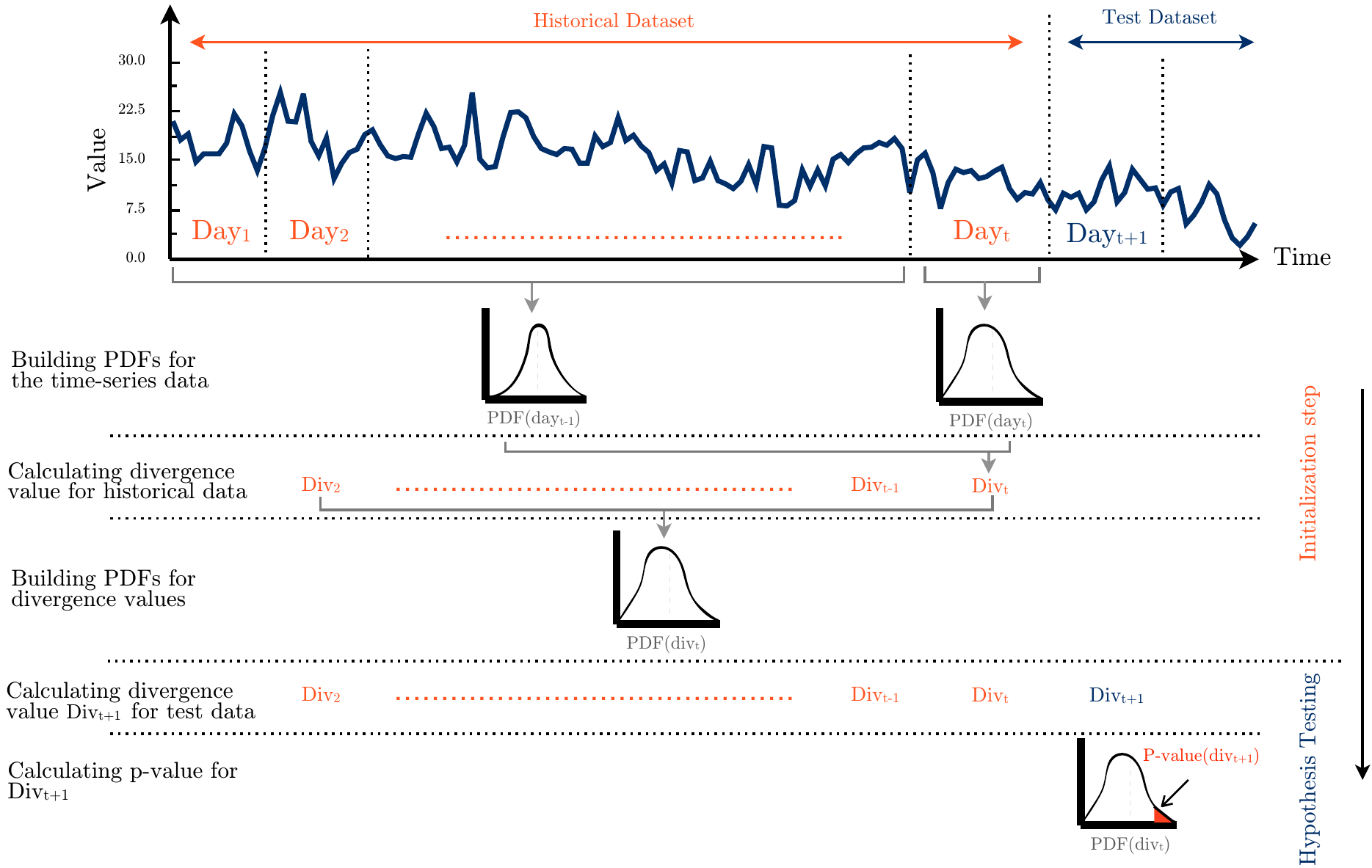}
\caption{Active detection method based on dynamic drift magnitude}
\label{fig:active_detection}
\end{figure*}

Algorithm \ref{algo:active} outlines our active drift adaptation approach. The algorithm first checks for drift occurrence and then makes predictions according to the drift analysis. We have developed the drift detection method based on monitoring the distribution of the divergence values calculated for each time window frame. The time window size is segmented according to the daily granularity as explained in Section \ref{sec:granularity}. As detailed in lines 3-6 of Algorithm \ref{algo:active} and shown in Fig. \ref{fig:active_detection}, after each time window frame, the divergence $Div_t = \textit{JSD}_{t}(P_{rf}\| P_{ts})$ is calculated between the probability density function of this specific day $d$ at time $t$ that represent the test samples $\textit{PDF}_{ts}$ and probability density function of historical data of all days that precede $t$ which represent the reference samples $\textit{PDF}_{rf}$, and so on. The divergence values are found using the Jensen-Shannon method (see Section \ref{sec:divergence}), and the divergence metric is calculated on the densities that are estimated using the \ac{kde} methods (see  Section \ref{sec:density}). For a training dataset of length $d$ periods that represent the temporal granularity of the drift detection mechanism, i.e., days in our case, $d-1$ divergence values are calculated that comprise the historical trends of changes in load consumption. These values will initialize the distribution of the divergence values before sliding on the evaluation dataset.

\begin{figure}
\centering
\includegraphics[width=0.48\textwidth]{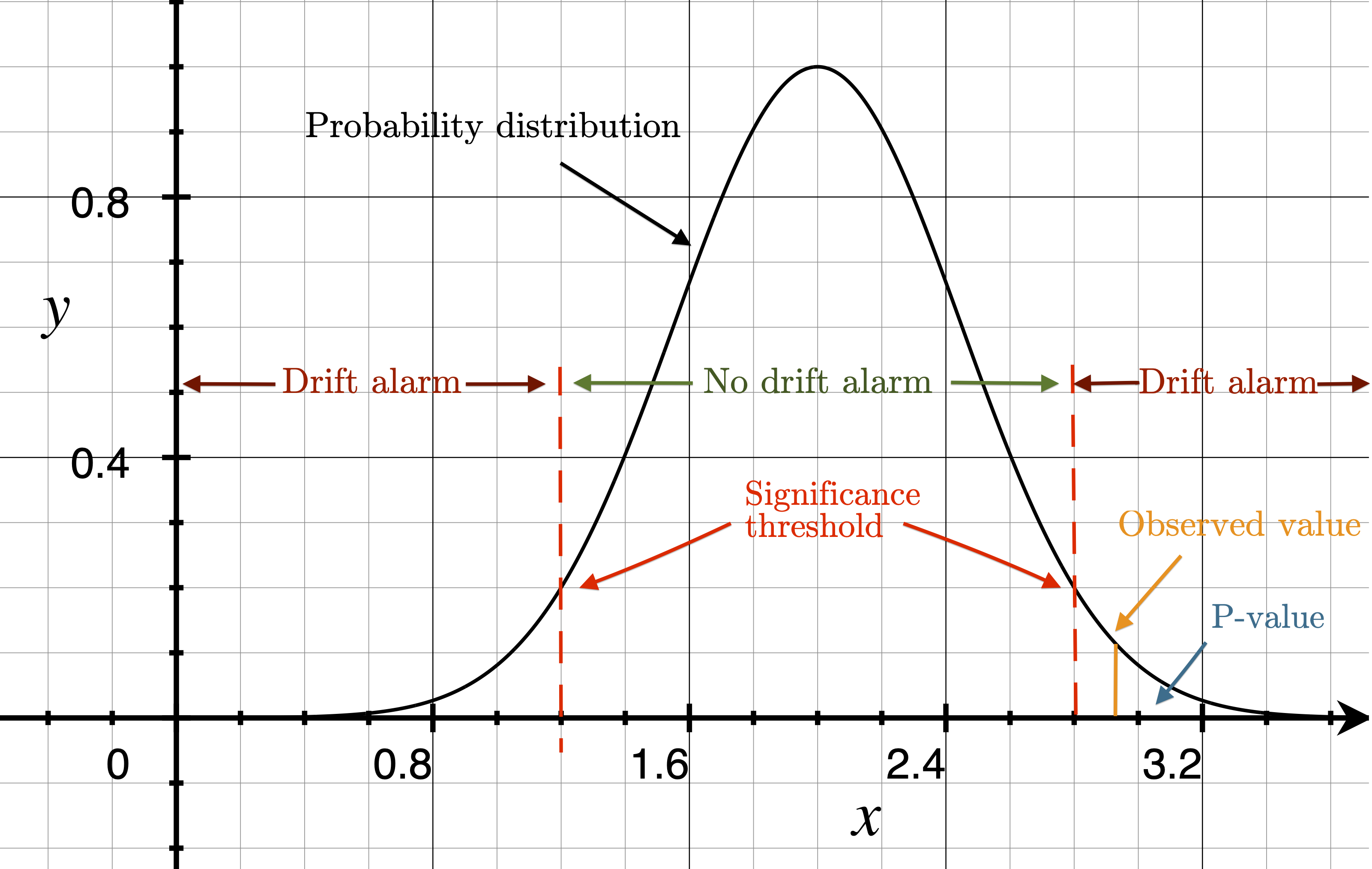}
\caption{Drift alarms based on hypothesis testing}
\label{fig:pvalue}
\end{figure}

After the initialization step, the divergence-based similarity score is found using the sliding window mechanism for each temporal granularity frame of the evaluation samples. Then, we test the null hypothesis of no change in the time-series data. The algorithm checks the p-value of the observed divergence value associated with the historical divergence value distribution. As illustrated in Fig. \ref{fig:pvalue}, the algorithm rejects the null hypothesis if the value exceeds the significance level $\tau$ and assumes a change in the behavior of the consumption load. The p-value is equal to the area under the curve of the probability density function \cite{knijnenburg2009fewer}. Our approach is dynamic for detecting the change point without fixing a detection threshold. This happens because the distribution of the divergence values evolves with the recording of more observations for load consumption. Furthermore, the significance level $\tau$ represents how extreme the divergence can be before signaling a change. The main advantage of using the p-value is that it is more persistent than a threshold that could be volatile for the change detection problem; hence, it is a more reliable indicator. The significance level $\tau$ controls the sensitivity to detect changes regardless of the magnitude of this change measured by the divergence metric.

\begin{algorithm}[t]
\caption{Active drift adaptation algorithm}
\begin{algorithmic}[1]
\Require Historical time-series samples $\mathbb{Y}(t-1)$, probability distribution of historical divergence values $\textit{PDF}(div_t)$, significance level $\tau$
\Ensure  Learner $f^{(t)}$, $PDF(div_{t+1})$
\State \textbf{Initialize}:   $f^{(t-1)}\big(\mathbb{Y}(t-1)\big)$
\For {$\mathbf{Y}(t)$, $t=2,\dots$}
\State Calculate $\textit{PDF}\big(\mathbf{Y}(t)\big) \leftarrow \textit{KDE}\big(\mathbf{Y}(t)\big)$
\State Calculate $\textit{PDF}\big(\mathbb{Y}(t-1)\big) \leftarrow \textit{KDE}\big(\mathbb{Y}(t-1)\big)$
\small \State Calculate $\textit{Div}_{t+1} \leftarrow  \textit{JSD} \Big(\textit{PDF}\big(\mathbf{Y}(t)\big) || \textit{PDF}\big(\mathbb{Y}(t-1)\big)\Big)$
\State Calculate $ \small \textit{PV}(div_{t+1}) \leftarrow \textit{P-value}\big(\textit{PDF}(div_t), \textit{Div}_{t+1} \big)$
\If{$\textit{PV}(div_{t+1}) < \tau$}
\State Update $f^{(t)} \leftarrow f^{(t-1)}(\mathbf{Y}(t))$
\State Predict $f^{(t)}(\mathbf{Y}(t+1))$
\Else
\State Predict $f^{(t-1)}(\mathbf{Y}(t+1))$
\EndIf
\State Calculate $E_{t+1}$
\EndFor
\State Update $\textit{PDF}(div_{t+1}) \leftarrow  [\textit{PDF}(div_t), div_{t+1}]$
\end{algorithmic}
\label{algo:active}
\end{algorithm}

\section{Experimental Evaluation and Results}
\label{results}
The performance of the ML predictions is evaluated using energy consumption records from real-world datasets that include nine households. However, for more detailed evaluations, we specifically selected two example households, namely Household 1 (an apartment) and Household 7 (a house). These households were chosen because they represent different types of households, allowing us to assess the performance of our model under varying conditions. The performance of ML predictions is affected by the detection of and reaction to drifts. Typically, drift adaptations improve the model's accuracy by performing a retraining or update process, which may impact the computational cost. Therefore, there is a trade-off between prediction accuracy and computational cost.

In this section, we answer the following questions:
\begin{itemize}
    \item \textbf{Prediction performance (Section \ref{sec:performance}):} How accurately do active and passive DA-LSTMs predict the energy demand for different parameters?
    
    \item \textbf{Computational Cost (Section \ref{sec:cost}):} What is the impact on the computational cost of both active and passive approaches?
    
    \item \textbf{Performance-cost trade-off (Section \ref{sec:tradeoff}):} What is the trade-off between performance and cost, and how can we use it to select the most suitable approach?
    
    \item \textbf{Real-life Pros and Cons (Section \ref{sec:discussion}):} What are the advantages and disadvantages of each approach in real-life problems?
\end{itemize}

\subsection{Consumption Dataset}
\label{sec:dataset}

To evaluate our approach, we used consumption data from the Swedish Energy Agency or Energimyndigheten \footnote{https://www.energimyndigheten.se/}. It contains the energy consumption of nine customers for one year at a resolution of ten minutes. Detailed information on the dataset is shown in Table \ref{tab: costumerinfo}. The table presents the start and end dates for the training and test dataset, divided by 75\% and 25\%, respectively. We chose this split since it showed a better prediction performance for datasets that are affected by seasonality \cite{jumin2021solar}. The validation dataset was created with $\frac{1}{6}$ of the entire training dataset for each customer, allowing for a reasonable amount of data to be used for training while still reserving a portion for validation and effectively capturing the complexity of the data. Additionally, information on the type of housing is given.

\begin{table*}[t]\centering
\caption{Information of energy consumption dataset}\label{tab: costumerinfo}
\setlength{\tabcolsep}{14.2pt}
\scriptsize
\begin{tabular}{lr|rrc|rrcr}\toprule
\multicolumn{2}{c|}{\textbf{Household information}} &\multicolumn{3}{c|}{\textbf{Training and validation dataset}} &\multicolumn{3}{c}{\textbf{Test dataset}} \\\cmidrule{1-8}
Household & Type & Start date & End date & Size &Start date &End date &Size \\\midrule
Household 1 &Apartment &18.08.2005 &24.06.2006 &44640 &24.06.2006 &23.08.2006 &8640 \\
Household 2 &Apartment &24.12.2005 &08.01.2007 &42768 &08.01.2007 &22.03.2007 &10512 \\
Household 3 &Apartment &22.03.2006 &29.01.2007 &44640 &29.01.2007 &30.03.2007 &8640 \\
Household 4 &Apartment &24.04.2007 &08.03.2008 &44640 &08.03.2008 &07.05.2008 &8640 \\
Household 5 &Apartment &21.08.2005 &05.07.2006 &44496 &05.07.2006 &04.09.2006 &8784 \\
Household 6 &Apartment &02.09.2005 &07.07.2006 &44784 &07.07.2006 &04.09.2006 &8496 \\
Household 7 &House &01.09.2005 &30.07.2006 &44064 &30.07.2006 &02.10.2006 &9216 \\
Household 8 &House &15.09.2005 &28.07.2006 &44496 &28.07.2006 &27.09.2006 &8784 \\
Household 9 &House &04.10.2006 &15.08.2007 &44496 &15.08.2007 &15.10.2007 &8784 \\
\bottomrule
\end{tabular}
\end{table*}

\subsection{Experimental setup}

\subsubsection{Baseline Methods}

To establish a comparable basis for the evaluation, we trained a baseline \ac{lstm} model and several baseline methods that are frequently used in load forecasting problems for each household individually. In particular, the baseline methods are: \ac{lstm}, Rolling \ac{arima}, Bagging Regression \cite{oza2001online} and an \ac{rnn} model \cite{fekri2021deep, Jagait2021Load}. For LSTM-based methods, the structure of the model consists of a singular \ac{lstm} layer combined with a dense layer. 
For each model, we individually tune the hyperparameters. The dataset of each household was divided into training and test sets according to the specifications mentioned in Section \ref{sec:dataset}. Each of the predictors takes 12 time steps (2 hours) as input and generates a prediction for the next six time steps, equal to an hour ahead. In addition, we created an \ac{lstm} model for the passive and active approaches, respectively.

\subsubsection{Evaluation Parameters}

For the active drift detection approach, three significance levels of p-values are defined: 0.07, 0.10, and 0.15. The significance level controls the extremeness of change magnitude for flagging a drift alarm. The higher the level value, the more probable the null hypotheses will be rejected because more values will fall within the drift-alarm region. The passive approach is equivalent to setting the p-value significance level to 1, meaning a drift always occurs. On the contrary, the baseline approach is equivalent to setting the p-value significance level to 0, i.e. a drift never occurs.
 

\subsubsection{Error Metrics}

To measure the prediction performance, we used \ac{mape} and \ac{rmse} as evaluation metrics since they are the most common error metrics used in the load forecasting literature according to a survey by Nti \textit{et al.} \cite{nti2020electricity}. To compare the results daily, we take the mean of 24 prediction errors as we detect changes on a daily basis. Since we predict an hour in advance, we calculate the mean of error metrics of 24 predictions for each day as follows:
\begin{equation}
\centering
\label{tab:mape_eq}
\text{MAPE} =\frac{1}{n} \sum_{t=1}^{n}\left|\frac{A_{t}-F_{t}}{A_{t}}\right| * 100,
\end{equation}

\begin{equation}
\centering
\text{RMSE}=\sqrt{\frac{1}{n} \sum_{t=1}^n\left(A_{t}-F_{t}\right)^2},
\end{equation}
where $A_{t}$ represents the vector of the actual consumed energy, $F_{t}$ the forecasted ones and $n$ is the number of predictions. 

\subsubsection{Computing Environments}
To gain insight into the operational efficiency of each approach, we deploy the solution on \ac{aws}\footnote{https://aws.amazon.com/} cloud service.  This will facilitate the trade-off analysis between the prediction performance and computational cost. \ac{aws} is one of the most popular public-cloud providers and frequently used as a deployment service for AI applications \cite{george2020usage}.
To quantify the computational cost, we measure the usage of \ac{cpu} and \ac{gpu} during run-time and the costs of running the approaches in a cloud environment (\ac{aws}). Table \ref{tab:specs} presents the specifications for the local machine and the corresponding \ac{aws} instance. We selected a G4 instance, namely: g4dn.4xlarge \footnote{ AWS instance: https://aws.amazon.com/ec2/instance-types/g4/}. For the calculation of the computational cost of the \ac{aws} instance, we measured the training time for the adaptation and calculated the on-demand price of the instance for that period, similar to \cite{pakdel2017adaptive}.

\begin{table}[b!]\centering
\caption{Computational specifications}\label{tab:specs}
\setlength{\tabcolsep}{20pt}
\scriptsize
\begin{tabular}{lrr}\toprule
&Local Environment \\\midrule
CPU &Intel Core i9-9900X CPU, 20 Threads \\
GPU &Nvidia GeForce RTX 2080, 11GB GDDR6 \\
RAM &64GB DDR4 \\\toprule
&AWS \\\midrule
CPU &Intel Xeon Scalable Processors, 16 Threads \\
GPU &Nvidia T4, 16GB GDDR6 \\
RAM &64GB DDR4 \\
\bottomrule
\end{tabular}
\end{table}


\subsection{Prediction Performance of DA-LSTM}
\label{sec:performance}

We conducted an exhaustive performance evaluation of our proposed DA-LSTM method against a conventional LSTM model, as it represents the main benchmark for comparison. Furthermore, we compared our method with several other baseline methods in the literature. However, we performed a more exhaustive evaluation for the conventional LSTM model, given that our proposed method is an improvement over it.

\subsubsection{Comparison Against Conventional LSTM}
\label{sec:base_comp}
\begin{figure}
    \centering
    \includegraphics[scale=0.6]{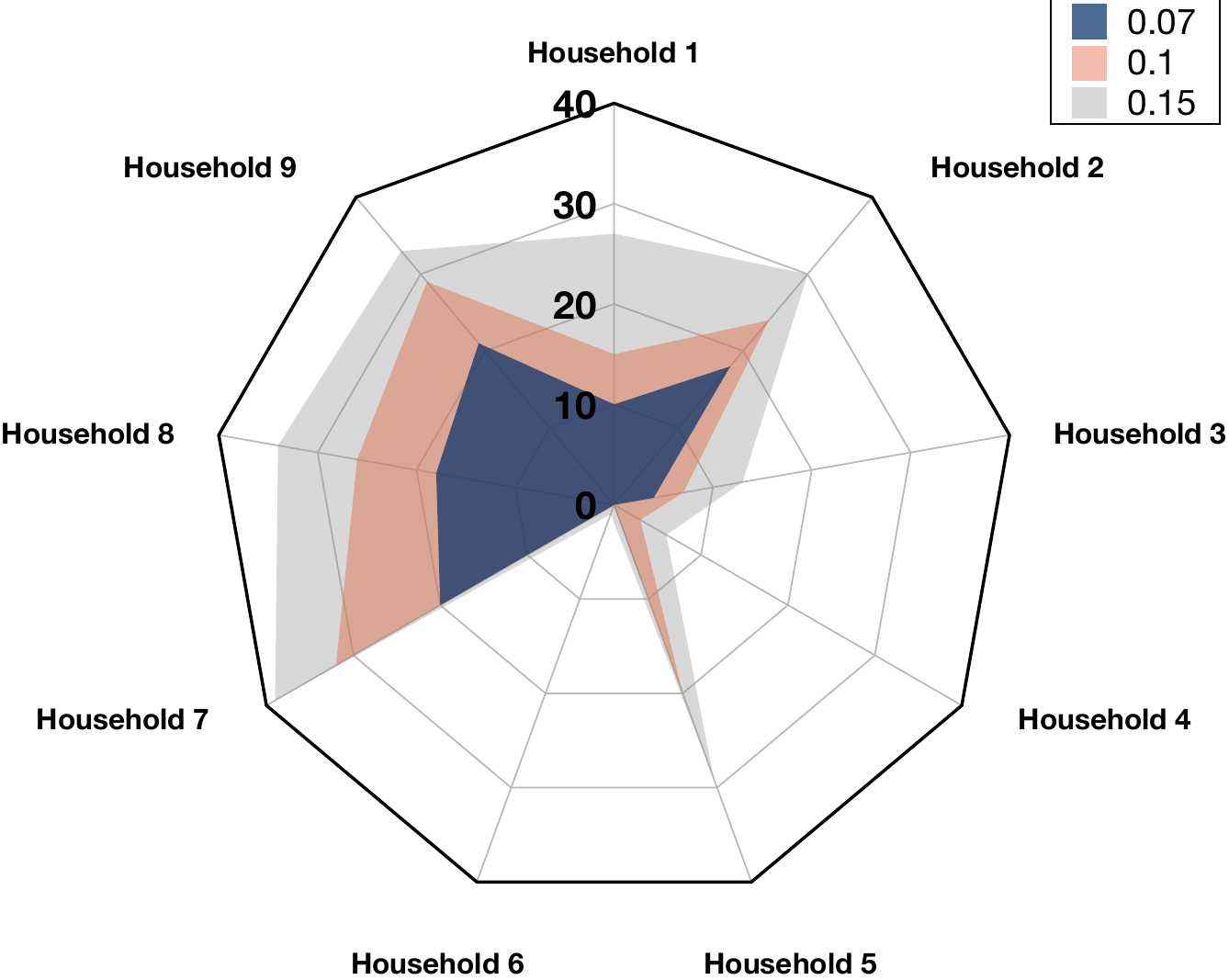}
    \caption{Detection counts per household for active approaches with different significance values $\tau$}
    \label{fig:detection_counts}
\end{figure}

The predictive performance of the different DA-LSTM strategies is evaluated and compared with the conventional baseline LSTM. We have also analyzed the drift detection counts for the active-based drift detection approach. The detection counts of all households with respect to the significance level value $\tau$ are summarized in Fig. \ref{fig:detection_counts}. We can see that households of the type \textit{House} have more irregular consumption patterns. Furthermore, some households do not have drift detected for a significance level of small values. This means that no adaptations are implemented for this household in the corresponding significance levels for the active approach.

Table \ref{tab:all_households_mape} shows the prediction error of all households observed for passive and active approaches for different parameters. For each household, we calculate the overall mean \ac{mape}, \ac{rmse} performance metrics, and the \ac{std} to check the variability of the errors during the evaluation period. In the case of active and passive approaches, an additional column represents the improvement (Imp) with respect to the prediction error reduction compared to the baseline model. For households without drift detection, see Fig. \ref{fig:detection_counts}, the baseline model is used without any adaptation.

Setting the significance level to a higher value increases the sensitivity to drift. This triggers more adaptation events, leading to an improvement in prediction quality (that is, a reduction in \ac{mape} and \ac{rmse}). From the Tables \ref{tab:all_households_mape} and \ref{tab:all_households_rmse}, we can see that the passive approach consistently outperforms all other approaches in all households. On average, the passive approach improves the \ac{mape} metric of all households produced by the baseline approach by 40.90~\%, and the \ac{rmse} metric by 50.16~\%. 

\begin{table*}
\centering
\caption{MAPE performance results of all households}
\label{tab:all_households_mape}
\setlength{\tabcolsep}{2pt}
\begin{tabular}{l|ll|lll|lll|lll|lll} 
\toprule
\multirow{2}{*}{\textbf{Household}} & \multicolumn{2}{c|}{\textbf{Baseline LSTM}} & \multicolumn{3}{c|}{\textbf{Active LSTM, $\tau=0.07$}}     & \multicolumn{3}{c|}{\textbf{Active LSTM, $\tau=0.1$}}      & \multicolumn{3}{c|}{\textbf{Active LSTM, $\tau=0.15$}}& \multicolumn{3}{c}{\textbf{Passive LSTM}}\\ 
& \textbf{Mean} & \textbf{STD}& \textbf{Mean} & \textbf{STD} & \textbf{Imp} & \textbf{Mean} & \textbf{STD} & \textbf{Imp} & \textbf{Mean} & \textbf{STD} & \textbf{Imp} & \textbf{Mean} & \textbf{STD} & \textbf{Imp}  \\
\midrule
\textbf{Household 1} &6.56 &1.10 &4.74 &0.87 &27.74~\% &3.70 &0.70 &43.60~\% &3.22 &0.68 &50.91~\% &2.95 &0.68 &55.03~\% \\
\textbf{Household 2} &4.05 &2.85 &4.03 &0.85 &0.49~\% &4.02 &0.79 &0.74~\% &3.50 &0.73 &13.58~\% &2.53 &0.16 &37.53~\% \\
\textbf{Household 3} &21.69 &13.43 &21.41 &9,11 &1.29~\% &21.23 &8.56 &2.12~\% &19.46 &7.98 &10.28~\% &11.24 &7.66 &48.18~\% \\
\textbf{Household 4} &9.48 &9.18 &9.48 &0.00 &0.00~\% &6.54 &0.93 &31.01~\% &5.55 &0.51 &41.46~\% &5.10 &0.18 &46.20~\% \\
\textbf{Household 5} &3.09 &3.46 &3.09 &0.00 &0.00~\% &2.76 &0.69 &10.68~\% &2.75 &0.53 &11.00~\% &2.66 &0.35 &13.92~\% \\
\textbf{Household 6} &4.34 &3.64 &4.34 &0.00 &0.00~\% &4.34 &0.00 &0.00~\% &3.59 &0.24 &17.28~\% &2.62 &0.06 &39.63~\% \\
\textbf{Household 7} &8.32 &2.08 &6.03 &1.62 &27.52~\% &4.80 &1.36 &42.31~\% &3.94 &1.16 &52.64~\% &4.09 &1.65 &50.84~\% \\
\textbf{Household 8} &6.11 &7.85 &6.01 &1.68 &1.64~\% &6.01 &1.45 &1.64~\% &5.43 &1.02&11.13~\% &5.33 &0.68 &12.77~\% \\
\textbf{Household 9} &8.34 &4.51 &7.25 &3.49 &13.07~\% &7.16 &2.98 &14.15~\% &6.96 &2.56 &16.55~\% &3.0 &0.54 &64.03~\% \\
\textbf{Average} &8.00 &5.34 &7.36 &2.04 &8.62~\% &6.73 &1.94 &16.25~\% &6.04 &1.71 &24.98~\% &4.39 &1.33 &40.90~\% \\
\bottomrule
\end{tabular}
\end{table*}

\begin{table*}
\centering
\caption{RMSE performance results of all households}
\label{tab:all_households_rmse}
\setlength{\tabcolsep}{2pt}
\begin{tabular}{l|ll|lll|lll|lll|lll} 
\toprule
\multirow{2}{*}{\textbf{Household}} & \multicolumn{2}{c|}{\textbf{Baseline LSTM}} & \multicolumn{3}{c|}{\textbf{Active LSTM, $\tau=0.07$}}     & \multicolumn{3}{c|}{\textbf{Active LSTM, $\tau=0.1$}}      & \multicolumn{3}{c|}{\textbf{Active LSTM, $\tau=0.15$}}& \multicolumn{3}{c}{\textbf{Passive LSTM}}\\ 
& \textbf{Mean} & \textbf{STD}& \textbf{Mean} & \textbf{STD} & \textbf{Imp} & \textbf{Mean} & \textbf{STD} & \textbf{Imp} & \textbf{Mean} & \textbf{STD} & \textbf{Imp} & \textbf{Mean} & \textbf{STD} & \textbf{Imp}  \\
\midrule
\textbf{Household 1} &0.45 &0.15 &0.32 &0.08 &28.88~\% &0.25 &0.07 &44.44~\% &0.23 &0.06 &48.88~\% &0.21 &0.07 &53.33~\% \\
\textbf{Household 2} &0.24 &0.17 &0.21 &0.07 &12.50~\% &0.18 &0.08 &25.00~\% &0.13 &0.07 &45.83~\% &0.10 &0.06 &58.33~\% \\
\textbf{Household 3} &0.74 &0.25 &0.52 &0.17 &29.72~\% &0.45 &0.15 &39.18~\% &0.39 &0.10 &47.29~\% &0.32 &0.08 &56.75~\% \\
\textbf{Household 4} &0.21 &0.11 &0.21 &0.00 &0.00~\% &0.15 &0.07 &28.57~\% &0.12 &0.07 &42.85~\% &0.11 &0.05 &47.61~\% \\
\textbf{Household 5} &0.19 &0.10 &0.19 &0.00 &0.00~\% &0.12 &0.08 &36.84~\% &0.10 &0.06 &47.36~\% &0.08 &0.06 &57.89~\% \\
\textbf{Household 6} &0.24 &0.14 &0.24 &0.00 &0.00~\% &0.24 &0.00 &0.00~\% &0.16 &0.05 &33.33~\% &0.15 &0.04 &37.50~\% \\

\textbf{Household 7} &0.42 &0.19 &0.25 &0.10 &40.47~\% &0.19 &0.08 &54.76~\% &0.18 &0.07 &55.89~\% &0.18 &0.06 &57.14~\% \\
\textbf{Household 8} &0.35 &0.16 &0.33 &0.12 &5.71~\% &0.28 &0.10 &19.99~\% &0.24 &0.09 &31.42~\% &0.20 &0.08 &42.85~\% \\
\textbf{Household 9} &0.30 &0.12 &0.26 &0.09 &13.33~\% &0.24 &0.08 &20.00~\% &0.21 &0.07 &30.00~\% &0.18 &0.07 &40.00~\% \\
\textbf{Average} &0.35 &0.15 &0.27 &0.10 &21.41~\% &0.23 &0.09 &32.18~\% &0.20 &0.07 &42.68~\% &0.17 &0.06 &50.16~\% \\
\bottomrule
\end{tabular}
\end{table*}

Specifically, the improvement in \ac{mape} reaches 64.03\% for Household 9, while it reaches 58.33\% for Household 2 on \ac{rmse}. However, for active approaches, the percentage of improvement increases with higher significance levels. The average percentage of improvement of \ac{mape} is 8.62~\% for the active approach with a significance level of $\tau=0.07$, reaching 24.98~\% for $\tau=0.15$. We observe a similar pattern of improvement in the \ac{rmse} metric, but with a larger effect size. For $\tau=0.07$ the improvement is 21.41~\% and reaches 42.68~\% for $\tau=0.15$.

\begin{figure*}
\centering

  \begin{subfigure}{\linewidth}
  \centering
  \includegraphics[scale=1]{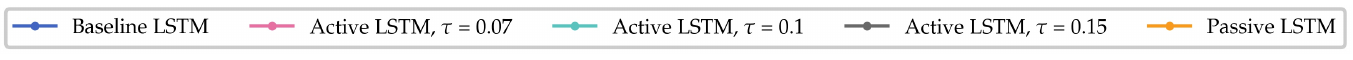}
  \end{subfigure}
  
\begin{subfigure}[t]{0.49\textwidth}
    \centering
    \includegraphics[scale=0.65]{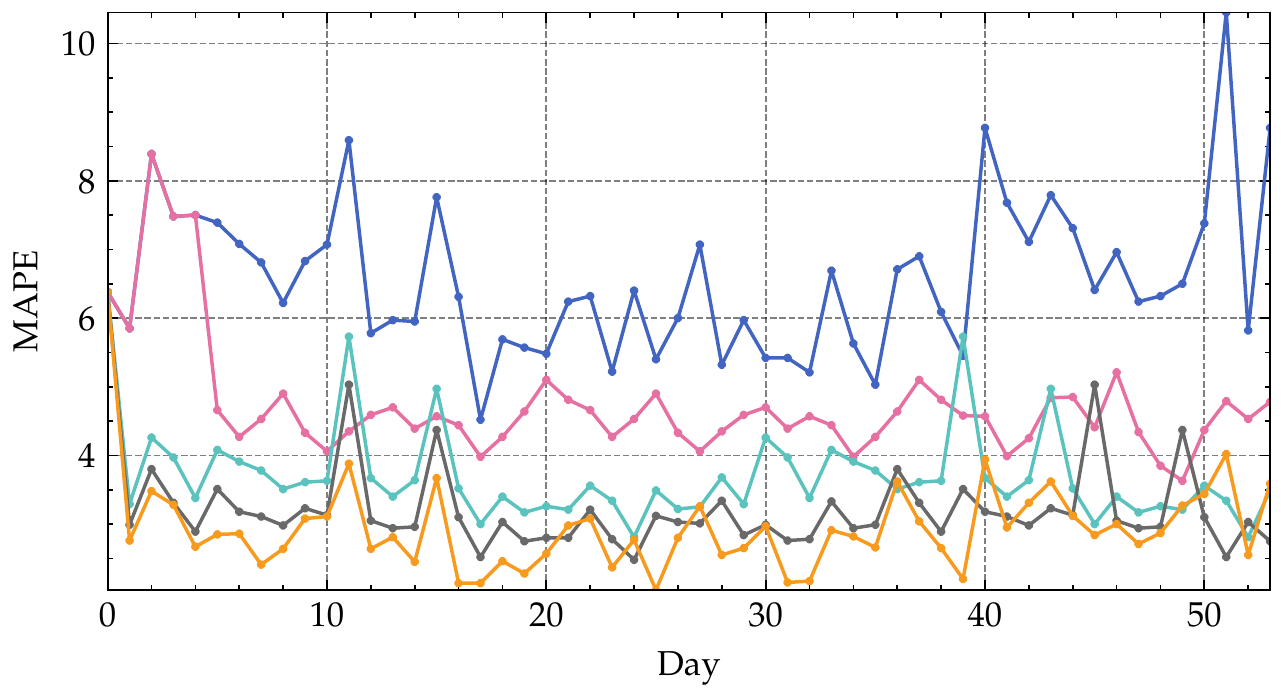}
    \caption{Overall MAPE for Household 1}
    \label{fig:customer1_mape}
\end{subfigure}
\hfill
\begin{subfigure}[t]{0.49\textwidth}
    \centering
    \includegraphics[scale=0.65]{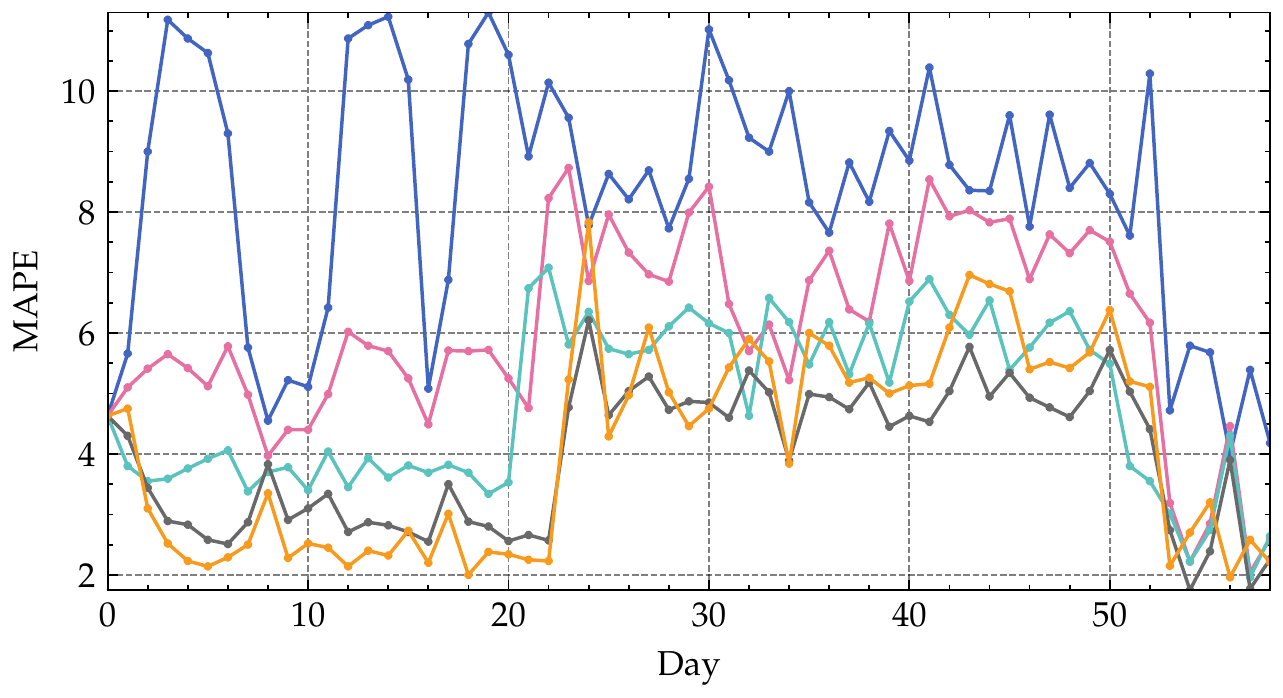}
    \caption{Overall MAPE for Household 7}
    \label{fig:customer7_mape}
\end{subfigure}

\vspace{0.3cm}

\begin{subfigure}[t]{0.49\textwidth}
    \centering
    \includegraphics[scale=0.65]{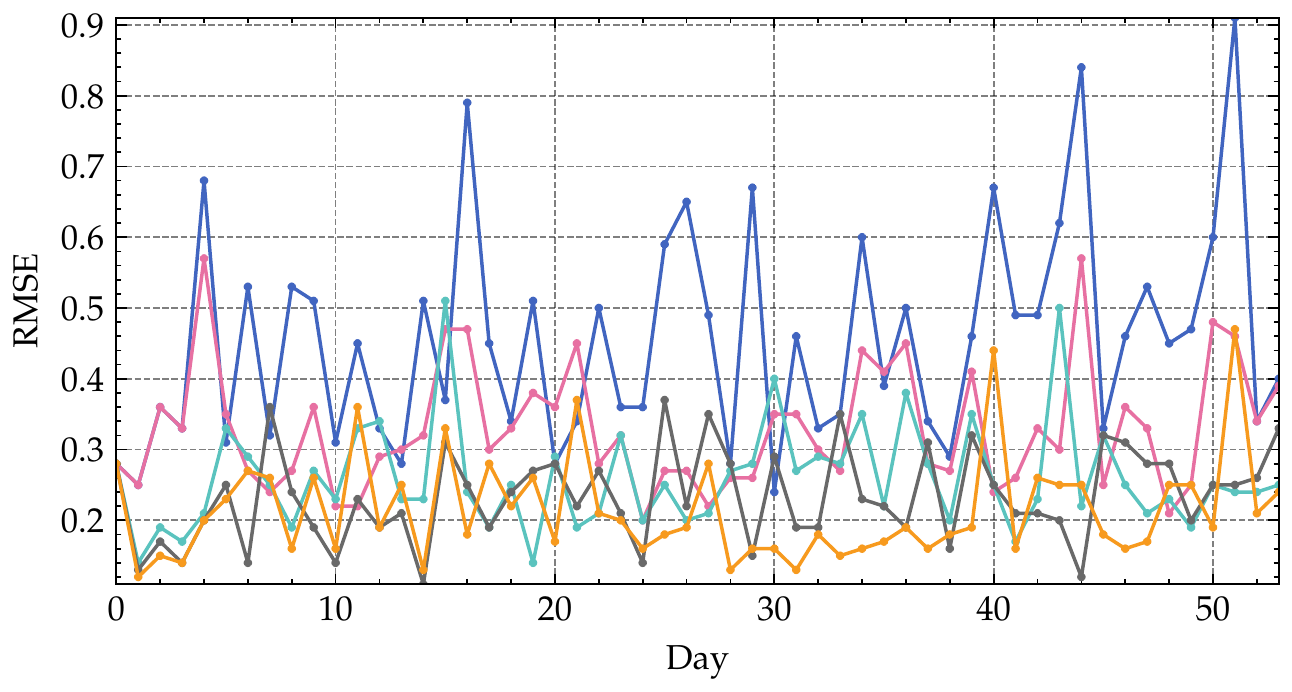}
    \caption{Overall RMSE for Household 1}
    \label{fig:customer1_rmse}
\end{subfigure}
\hfill
\begin{subfigure}[t]{0.49\textwidth}
    \centering
    \includegraphics[scale=0.65]{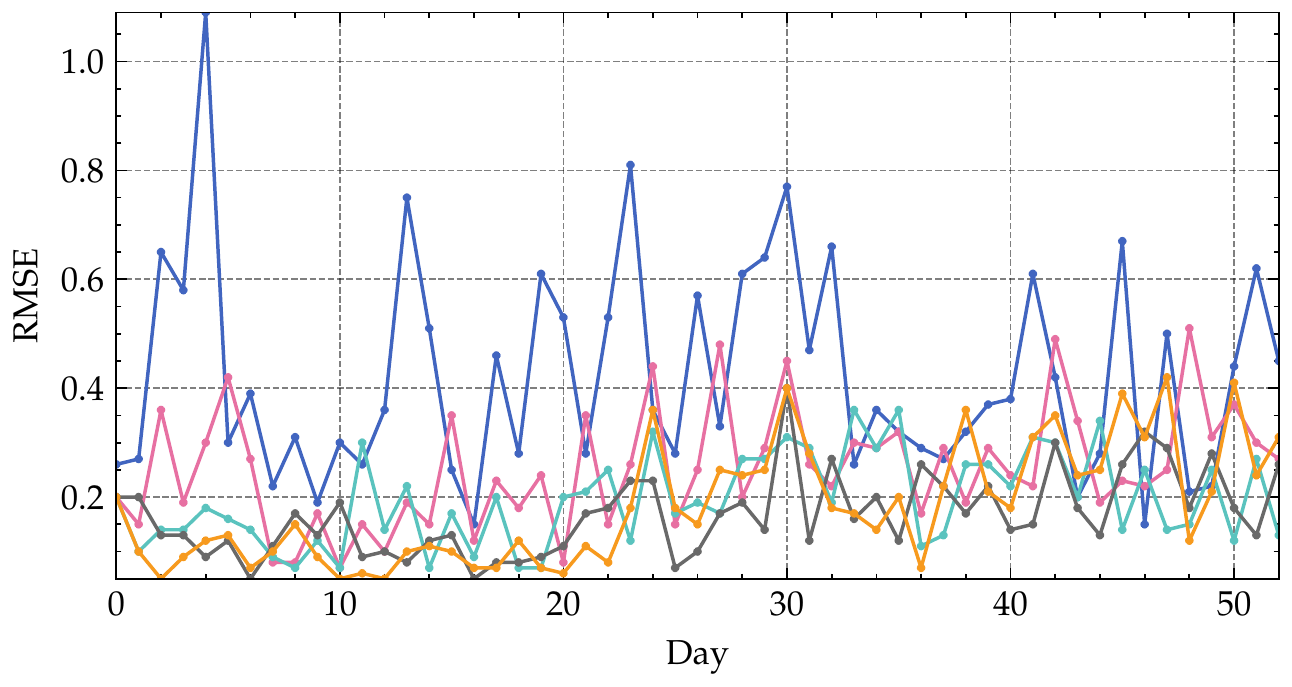}
    \caption{Overall RMSE for Household 7}
    \label{fig:customer7_rmse}
\end{subfigure}

\caption{Performance evaluation for LSTM-based methods}
\label{fig:per_metrics}

\end{figure*}

To evaluate performance on a more granular level, we plotted the daily performance metrics for two examples of each type of household. Fig. \ref{fig:customer1_mape} and Fig. \ref{fig:customer1_rmse} present the daily \ac{mape} and \ac{rmse} for Household 1, respectively. While Fig. \ref{fig:customer7_mape} and Fig. \ref{fig:customer7_rmse} present the performance metrics for Household 7. As we can see, for both households, the baseline model has the highest error rates compared to when using the drift-adaptive  approaches for each day of the test data. Moreover, the error curves for DA-LSTM approaches are stacked over most days in both figures. However, in a few cases, the active approach performs better than the passive approach due to the non-determinism property of the output of neural networks \cite{taylor2003verification}.

\subsubsection{Comparison Against Additional Baseline Methods from the Literature}
\label{sec:add_base_comp}
\begin{figure*}
\label{fig:Baseline}
\centering
  \begin{subfigure}{\linewidth}
  \centering
  \includegraphics[scale=0.48]{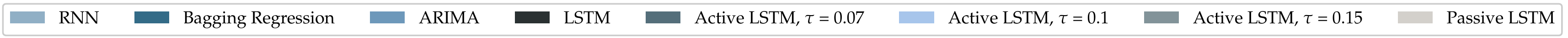}
  \end{subfigure}
  
\begin{subfigure}[t]{0.49\textwidth}
    \centering
    \includegraphics[scale=0.64]{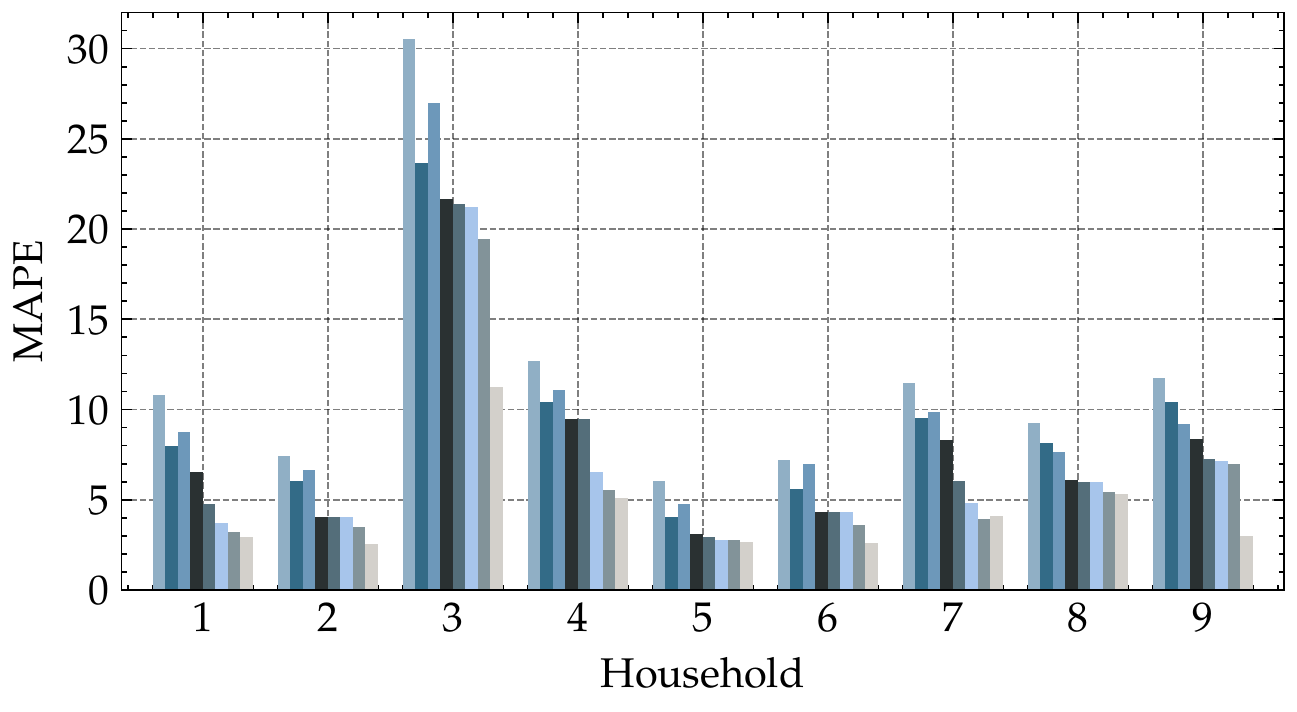}
    \caption{Overall MAPE for Household 1}
    \label{fig:customer_MAPE_Base}
\end{subfigure}
\hfill
\begin{subfigure}[t]{0.49\textwidth}
    \centering
    \includegraphics[scale=0.64]{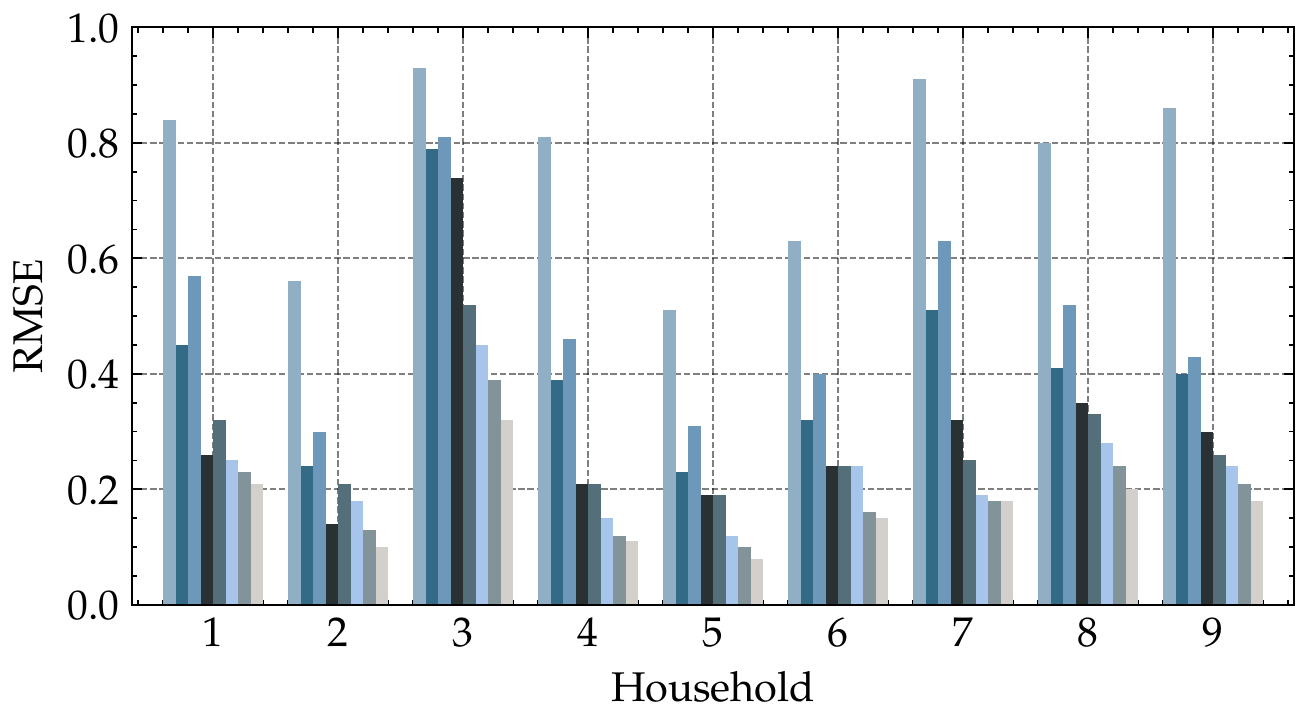}
    \caption{Overall RMSE for Household 7}
    \label{fig:customer_RMSE_Base}
\end{subfigure}

\caption{Performance evaluation for baseline and DA-LSTM methods}
\label{fig:per_metrics1}

\end{figure*}

To further verify the effectiveness of our proposed method, we performed additional evaluations against several commonly used baseline methods in the literature. Fig. \ref{fig:customer_MAPE_Base} and \ref{fig:customer_RMSE_Base} show the performance of the baseline methods and DA-LSTM on the MAPE and RMSE metrics. Regarding the non-LSTM-based baseline methods, the Bagging Regression method performs better than the RNN and ARIMA methods. Specifically, the performance of the Bagging Regression model is the most similar to the LSTM baseline model. Overall, the results show that the proposed DA-LSTM method outperforms all other baseline methods in all cases by a considerable margin for MAPE and RMSE for all households. DA-LSTM achieved a maximum improvement of 62.8\% over \ac{rnn} on \ac{mape} and 84.51\% on \ac{rmse}.

\subsubsection{DA-LSTM Hyperparameter Optimization Results}

As described in Section \ref{sec:hpo}, we performed the \ac{hpo} for all of the \ac{lstm} models to find the optimal parameters. Tables 6 to 9 illustrate the outcomes of hyperparameter tuning using validation data. Specifically, to provide a glimpse into the range of values that may be effective for this approach, examples of the selected hyperparameter values are reported in Tables \ref{tab:hpotable_1_07} and \ref{tab:hpotable_1_10} for Household 1, and Tables \ref{tab:hpotable_7_07} and \ref{tab:hpotable_7_10} for Household 7, with significance levels $\tau=0.07$ and $\tau=0.10$, respectively. In contrast to the results reported in Tables \ref{tab:all_households_mape} and \ref{tab:all_households_rmse}, which present the evaluation results for Household 1 and 7 respectively and are aggregated for the entire test dataset of each household. The tables represent the best combination of hyperparameters that resulted in the optimal performance in terms of loss value (MAPE). An interesting observation is that some hyperparameters, such as LR and DR, appear to have similar values across different detection numbers. For example, in Table \ref{tab:hpotable_7_07}, detection numbers 1, 6, 10, and 16 all have the same LR and DR values of 0.001 and 0.4, respectively. Similarly, in Table \ref{tab:hpotable_7_10}, detection numbers 1, 9, 17, and 23 have the same LR and DR values. On the contrary, the parameter $N_U$ exhibits greater variability in different detection numbers than the other hyperparameters, suggesting that it is more sensitive to data changes.

\begin{table}[!htp]\centering
\caption{DA-LSTM Hyperparameter values for Household 1, $\tau=0.07$}\label{tab:hpotable_1_07}
\scriptsize
\setlength{\tabcolsep}{9pt}

\begin{tabular}{crrrr}\toprule
Detection Number &$LR$ &$DR$ &$N_{U}$ &Loss
\\\midrule
1 &0.001 &0.4 &192 &3.13\\
2 &0.0001 &0 &160 & 3.14\\
3 &0.001 &0.3 &160 & 3.12\\
4 &0.001 &0.3 &192 & 3.10\\
5 &0.0001 &0 &128 & 3.10\\
6 &0.0001 &0 &256 & 3.08\\
7 &0.001 &0.2 &224 & 3.07\\
8 &0.001 &0.1 &256 & 3.05\\
9 &0.0001 &0 &192 & 3.05\\
10 &0.001 &0.4 &192 & 3.00\\
\bottomrule
\end{tabular}
\end{table}

\begin{table}[!htp]\centering
\caption{DA-LSTM Hyperparameter values for Household 7, $\tau=0.07$}\label{tab:hpotable_7_07}
\scriptsize
\setlength{\tabcolsep}{9pt}

\begin{tabular}{crrrr}\toprule
Detection Number &$LR$ &$DR$ &$N_{U}$ &Loss
\\\midrule
1 &0.001 &0.4 &224 &2.74\\
2 &0.001 &0 &288 & 2.76\\
3 &0.0001 &0 &224 & 2.75\\
4 &0.001 &0.2 &192 & 2.75\\
5 &0.0001 &0 &224 & 2.75\\
6 &0.001 &0.4 &320 & 2.75\\
7 &0.0001 &0 &320 & 2.75\\
8 &0.001 &0.3 &224 & 2.77\\
9 &0.0001 &0 &480 & 2.79\\
10 &0.001 &0.4 &224 & 2.79\\
11 &0.0001 &0.1 &384 & 2.81\\
12 &0.0001 &0.2 &320 & 2.84\\
13 &0.001 &0 &480 & 2.83\\
14 &0.001 &0.3 &224 & 2.82 \\
15 &0.0001 &0.1 &416 & 2.83\\
16 &0.001 &0.4 &320 & 2.83\\
17 &0.001 &0 &320 & 2.85\\
18 &0.0001 &0 &224 & 2.88\\
19 &0.01 &0 &256 & 2.88\\
20 &0.001 &0.2 &256 & 2.87\\
\bottomrule
\end{tabular}
\end{table}

\begin{table}[!htp]\centering
\caption{DA-LSTM Hyperparameter values for Household 1, $\tau=0.10$}\label{tab:hpotable_1_10}
\scriptsize
\setlength{\tabcolsep}{9pt}

\begin{tabular}{crrrr}\toprule
Detection Number &$LR$ &$DR$ &$N_{U}$ & Loss\\\midrule
1 &0.01 &0 &96 & 3.17 \\
2 &0.001 &0.4 &256 & 3.24\\
3 &0.001 &0 &288 & 3.14\\
4 &0.001 &0.3 &288 & 3.10\\
5 &0.0001 &0.1 &224 & 3.40\\
6 &0.0001 &0.2 &224 & 3.79\\
7 &0.0001 &0.1 &224 & 3.24\\
8 &0.001 &0.1 &288 & 3.69 \\
9 &0.001 &0.2 &256 & 3.08\\
10 &0.001 &0 &288 & 3.12\\
11 &0.0001 &0.1 &256 & 3.18\\
12 &0.0001 &0 &224 & 3.68\\
13 &0.001 &0.4 &352 & 2.99\\
14 &0.001 &0.1 &192 & 3.00\\
15 &0.0001 &0 &288 & 2.98\\
\bottomrule
\end{tabular}
\end{table}

\begin{table}[!htp]\centering
\caption{DA-LSTM Hyperparameter values for Household 7, $\tau=0.10$}\label{tab:hpotable_7_10}
\scriptsize
\setlength{\tabcolsep}{9pt}

\begin{tabular}{crrrr}\toprule
Detection Number &$LR$ &$DR$ &$N_{U}$ & Loss\\\midrule
1 &0.001 &0.4 &224 & 4.36 \\
2 &0.001 &0 &288 & 2.74\\
3 &0.0001 &0 &224 & 2.74\\
4 &0.0001 &0 &288 & 2.75\\
5 &0.001 &0.2 &192 & 3.47\\
6 &0.001 &0 &224 & 2.77\\
7 &0.0001 &0 &224 & 2.74\\
8 &0.001 &0.2 &480 & 2.76 \\
9 &0.001 &0.4 &320 & 2.76\\
10 &0.001 &0 &448 & 2.77\\
11 &0.0001 &0 &320 & 2.78\\
12 &0.001 &0 &224 & 2.79\\
13 &0.001 &0 &480 & 3.16\\
14 &0.001 &0 &448 & 2.82\\
15 &0.001 &0.3 &224 & 4.29\\
16 &0.0001 &0 &480 & 2.83\\
17 &0.001 &0.4 &224 & 2.83\\
18 &0.0001 &0.1 &384 & 2.84\\
19 &0.0001 &0.2 &320 & 2.83\\
20 &0.001 &0 &480 & 2.83\\
21 &0.001 &0.3 &224 & 2.84\\
22 &0.0001 &0.1 &416 & 2.85\\
23 &0.001 &0.4 &320 & 2.87\\
24 &0.001 &0 &256 & 2.89\\
25 &0.0001 &0 &352 & 2.88\\
26 &0.0001 &0 &288 & 2.88\\
27 &0.0001 &0 &352 & 2.89\\
28 &0.0001 &0 &128 & 2.88\\
29 &0.001 &0 &320 & 2.89\\
30 &0.0001 &0 &224 & 2.89\\
31 &0.01 &0 &256 & 2.87\\
32 &0.001 &0.2 &256 & 2.87\\
\bottomrule
\end{tabular}
\end{table}

\subsubsection{Learning Curves for LSTM-Based Methods}
\begin{figure*}
\centering
  \begin{subfigure}{\linewidth}
  \centering
  \includegraphics[scale=1.2]{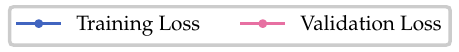}
  \end{subfigure}
  
\begin{subfigure}[t]{0.49\textwidth}
    \centering
    \includegraphics[scale=0.65]{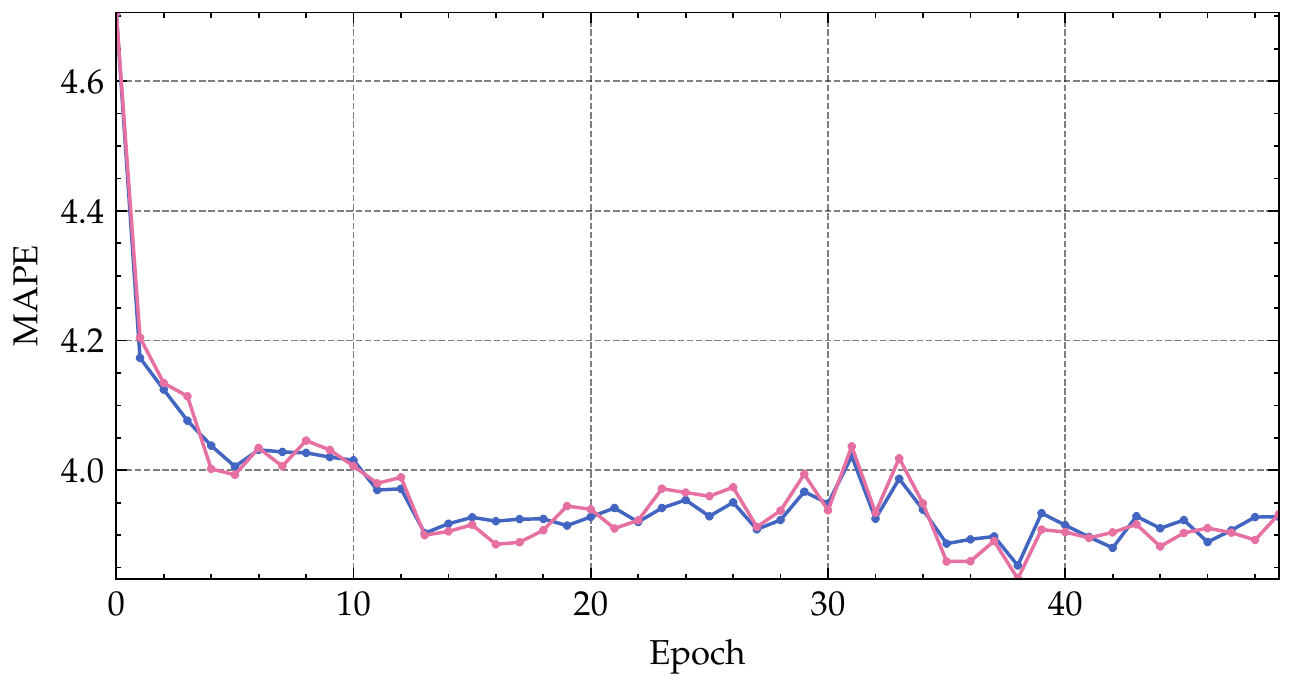}
    \caption{Training and Validation curves for Household 1}
    \label{fig:customer1_fit}
\end{subfigure}
\hfill
\begin{subfigure}[t]{0.49\textwidth}
    \centering
    \includegraphics[scale=0.65]{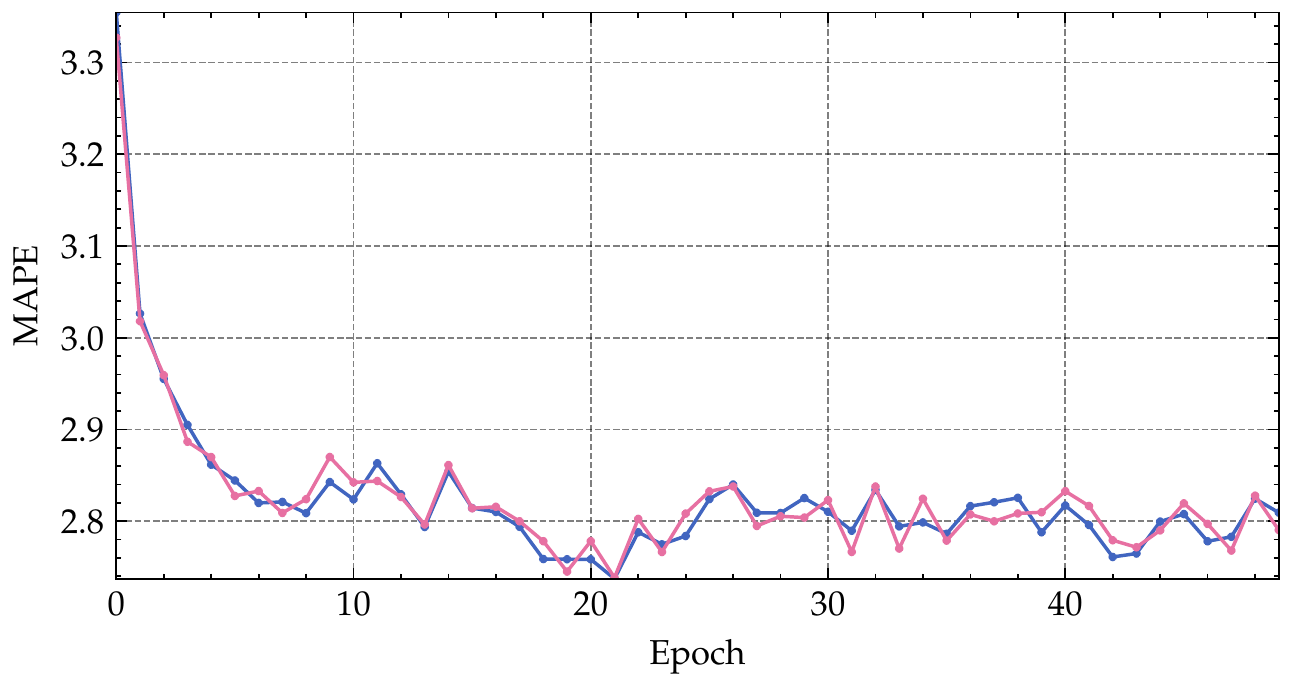}
    \caption{Training and Validation curves for Household 7}
    \label{fig:customer7_train_val}
\end{subfigure}

\caption{Training and validation curves}
\label{fig:train_val}
\end{figure*}

To corroborate the performance during the training process, we analyze the learning curves to inspect whether the LSTM-based models are overfitting or underfitting. Figs. \ref{fig:customer1_fit} and \ref{fig:customer7_train_val} illustrate the training and validation curves for Household 1 and Household 7, respectively. The curves show that the model can capture the patterns in the training data without overemphasizing noise, while still being able to generalize to new data as seen in the validation set and there is no specific sign of over- or under-fitting.

\subsection{Computational Cost}
\label{sec:cost}
\begin{figure*}
    \centering
    \includegraphics[scale=1]{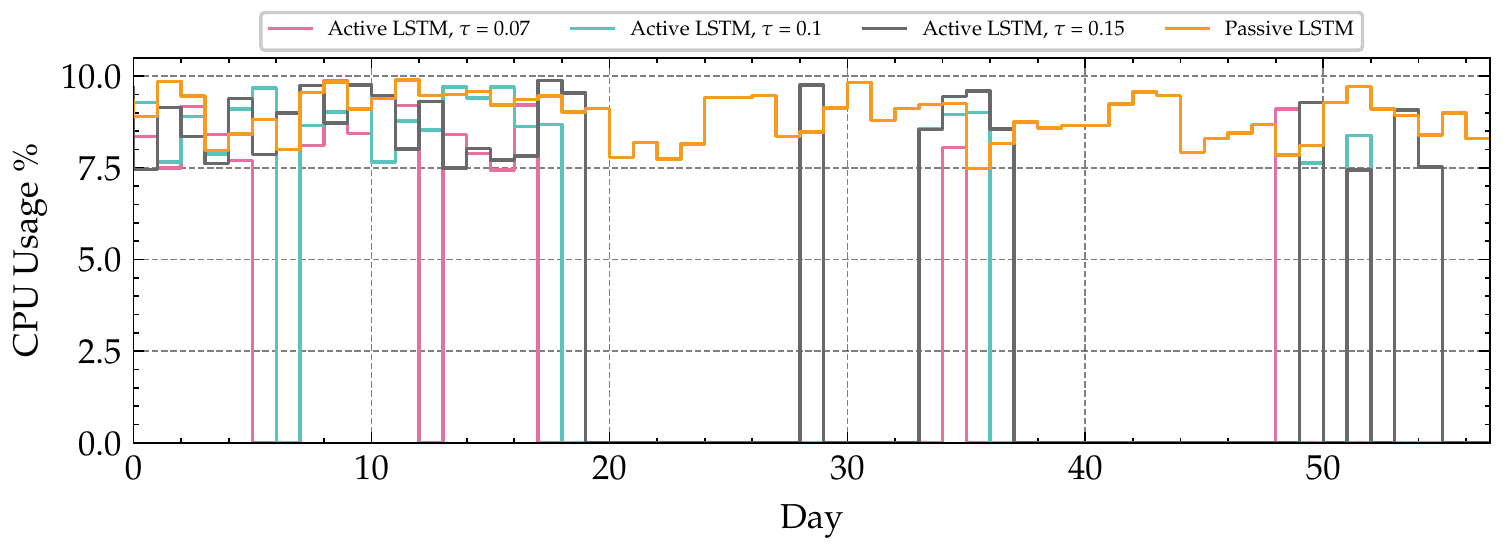}
    \caption{Daily CPU usage percentage of forecasting model for Household 7}
    \label{fig:cpu}
\end{figure*}

\begin{figure*}
    \centering
    \includegraphics[scale=1]{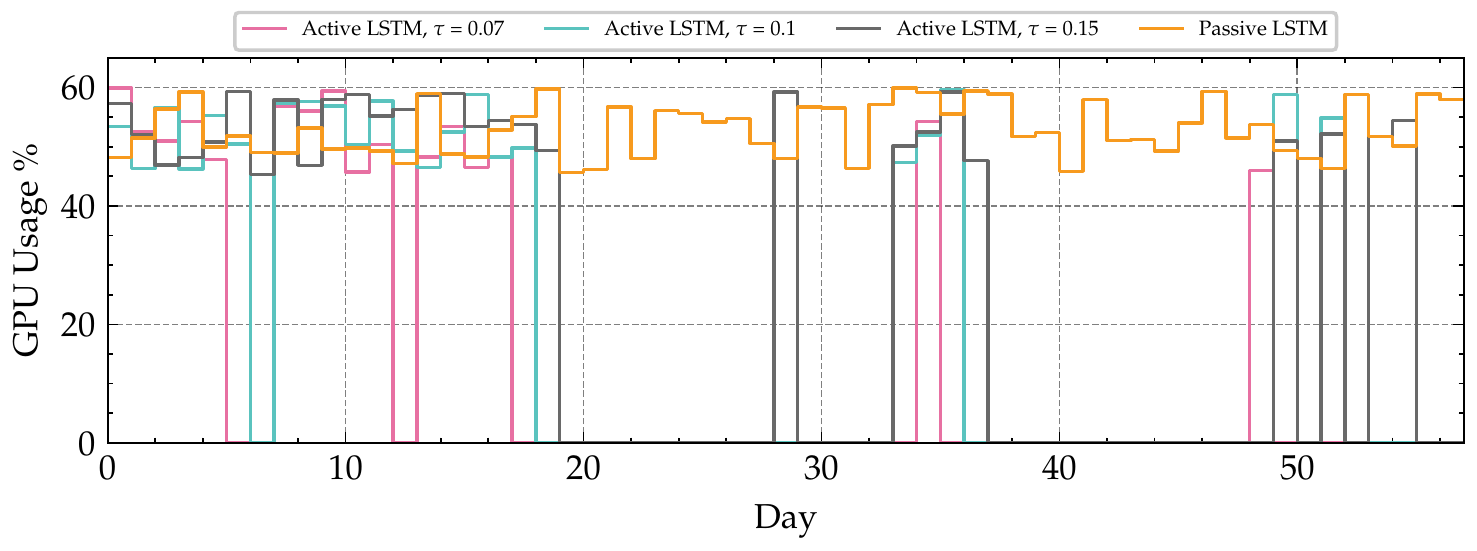}
    \caption{Daily GPU usage percentage of forecasting model for Household 7}
    \label{fig:gpu}
\end{figure*}

Adjusting the models to adapt to changes is associated with additional computational costs.  Figs. \ref{fig:cpu} and \ref{fig:gpu} present the \ac{cpu} and \ac{gpu} usage for each of the active and passive approaches of Household 7; the household with the highest number of detected drifts. As can be seen, when the adaptation signal is activated due to a change in behavior, we have a similar resource utilization pattern in both approaches. For \ac{cpu}, it fluctuates between 7\% and 10\% per thread, while for \ac{gpu}, the utilization of resources is between 45\% and 60\%. 
The conventional \ac{lstm} model, as described in Section \ref{sec:convlstm}, only computes sequentially. But both TensorFlow and PyTorch introduced a cuDNNLSTM wrapper \footnote{CUDNNLSTM: https://www.tensorflow.org/api\_docs/python/tf/compat/v1/keras/layers/CuDNNLSTM} for the normal \ac{lstm} implementation, which enables its partly parallelization \cite{braun2018lstm, appleyard2016optimizing}

\begin{figure*}
    \centering
    \includegraphics[scale=1]{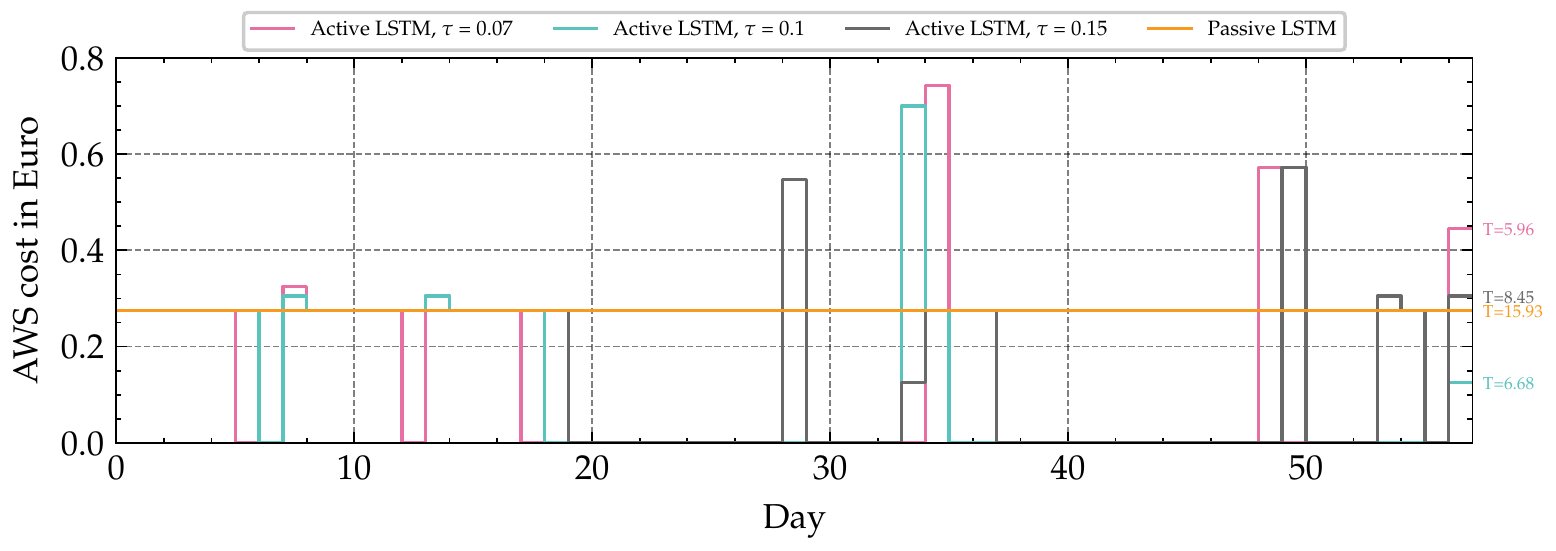}
    \caption{Daily AWS cost of forecasting model for Household 7, T is the total cost of the entire period}
    \label{fig:aws}
\end{figure*}

In Table \ref{tab:overallcost}, the total \ac{aws} costs in Euros for the entire evaluation period of each household are presented.  The cost is calculated based on the number of minutes needed for the adaptation with the on-demand pricing of the instance. The table shows that the passive approach is at least twice as costly as the active approach with the selected significance levels. On the daily level, Fig. \ref{fig:aws} shows the computational cost for Household 7 to adapt the model on the \ac{aws} instance. As can be seen,  the passive approach shows a constant cost throughout the whole observed evaluation period. This is because the training time and hyperparameter tuning for each daily period require the same timeframe. However, the active approaches impose a variable cost. For example, when $\tau=0.07$, a significant change in computational cost can be observed on day 37 with a cost of 0.744 for the training in ML adaptation. This increase in cost occurs when the time between changes is prolonged, leading to a more extensive training dataset for the adaptation of \ac{lstm} and, consequently, to higher costs. The total cost of running the system for 58 consecutive days for the passive approach is 15.93 Euros, which corresponds to 0.27 per day. In comparison, the cost for the same period when using the active approaches are: 5.96, 6.67, and 8.44 with an increasing $\tau$, respectively. It is visible that even with a single household over the time span of only 58 days, the proposed approach can save about 53\% of the computational cost with $\tau= 0.15$. Scaling the number of households in a typical case of only one low voltage feeder of tens of households, e.g., 70 households, the savings in practice are even much higher in the case of several low voltage feeders on a residential region of some hundreds of residencies. Thus, this method would result in high savings in terms of costs.

\begin{table}[!htp]\centering
\caption{Total computational cost per household in EUR }\label{tab:overallcost}
\setlength{\tabcolsep}{8pt}
\scriptsize
\begin{tabular}{lrrrrr}\toprule
Household&$\tau=0.07$ &$\tau=0.10$ &$\tau=0.15$ &Passive \\\midrule
Household 1 &7.53 &8.86 &9.88 &24.17 \\
Household 2 &7.44 &8.11 &11.02 &31.31 \\
Household 3 &4.56 &5.50 &6.50 &26.37 \\
Household 4 &0.00 &4.10 &4.91 &22.80 \\
Household 5 &0.00 &7.66 &9.89 &18.95 \\
Household 6 &0.00 &0.00 &2.52 &20.60 \\
Household 7 &5.96 &6.68 &8.45 &15.93 \\
Household 8 &9.42 &10.56 &12.56 &31.04 \\
Household 9 &7.27 &9.25 &10.47 &20.60 \\
\bottomrule
\end{tabular}
\end{table}

\subsection{Performance-Cost Trade-off}
\label{sec:tradeoff}
In the previous section, it was evident that there is a trade-off between predictive performance and the associated computational cost of each approach. In most cases, the passive approach is the most powerful approach in terms of prediction accuracy. However, at the same time, it is the most expensive approach in terms of computational costs. Similarly, the active approach showed behaviors identical to those of the passive approach. Effective approaches are associated with additional workload costs due to the number of adaptation events. We have calculated a metric that combines predictive performance and computational costs to further analyze the trade-off. 

The simplest measure that combines both performance and costs is the ratio between the two metrics \cite{vandierendonck2017comparison}. The trade-off score (TS) can be simply calculated as follows:
\begin{equation}
\mathrm{TS}=\frac{\mathrm{Performance}}{\mathrm{Cost}},
\label{eq:tradeoff}
\end{equation}
where $\mathrm{Performance}$ is the metric that is used to evaluate the predictive performance of the approach, we chose the MAPE metric as it yielded similar results to RMSE, and $\mathrm{cost}$ is the total cost of the associated \ac{aws} instance for the evaluation period.  The most efficient approach is one that maximizes performance per unit cost ratio (TS). Table \ref{tab:ratio} presents the results in the household of computing the trade-off metric for each approach. The metric is calculated based on the improvement of the error rate (shown in Table \ref{tab:all_households_mape}) and the total cost of the \ac{aws} instance (shown in Table \ref{tab:overallcost}).

\begin{table}[!htp]\centering
\caption{Trade-off metric between performance improvement and cost}\label{tab:ratio}
\setlength{\tabcolsep}{8pt}
\scriptsize
\begin{tabular}{lrrrrr}\toprule
Household&$\tau=0.07$ &$\tau=0.10$ &$\tau=0.15$ &Passive \\\midrule
Household 1 &3.68 &4.92 & \textbf{5.15*}&2.28 \\
Household 2 &0.07 &0.09 &\textbf{1.23*} &1.2 \\
Household 3 &0.28 &0.39 &1.58 &\textbf{1.83*} \\
Household 4 &0.00 &7.57 &\textbf{8.44*} &2.03 \\
Household 5 &0.00 &\textbf{1.39*} &1.11 &0.73 \\
Household 6 &0.00 &0.00 &\textbf{6.87*} &1.92 \\
Household 7 &4.62 &\textbf{6.34*} &6.23 &3.19 \\
Household 8 &0.17 &0.16 &\textbf{0.89*} &0.41 \\
Household 9 &1.88 &1.53 &1.58 &\textbf{3.11*} \\
\bottomrule
\end{tabular}
\end{table}

We can see that there is no single approach that fits all household data. However, the results show that the active strategy with a significance level of $\tau=0.15$ is the most efficient approach in terms of the TS metric for five out of nine households. Meanwhile, despite its low error rate, the passive strategy is the most effective approach for two households only. This means that the additional costs do not move proportionally with the performance improvement rates.  Furthermore, the active strategy with a significance level of $\tau=0.1$ has been the most effective approach for two households in total.

\subsection{Discussion}
\label{sec:discussion}
In this section, we presented a detailed experimental evaluation of the performance of DA-LSTM using a real-life dataset. Our DA-LSTM approach introduces a novel adaptive mechanism that dynamically adjusts the LSTM model to changing data patterns, allowing for accurate and robust predictions in the presence of concept drift. We employed an active-based drift detection approach, which leverages statistical significance testing, to detect drift events and trigger model adaptations accordingly. This active approach is a key contribution of our work, as it enables the LSTM model to dynamically adapt to changes and maintain high prediction performance over time.  Additionally, compared to other drift adaptive methods in interval load forecasting such as \cite{fekri2021deep, Jagait2021Load, Ji2021Enhancing}, which require setting a threshold for drift magnitude, the dynamic drift detection methodology implemented in the proposed method is a flexible tool for detecting changes in consumption patterns without pre-defining a threshold. As it can be challenging to determine an appropriate threshold for drift detection in real-life settings. This feature adds to the practicality and usefulness of the proposed method in real-world applications. Moreover, the incremental learning technique employed in our DA-LSTM approach is advantageous as it builds on the previously learned knowledge and avoids the shortcomings of complete forgetting, as required in some studies that only rely on the most recent observations as a prerequisite \cite{fenza2019drift, chitalia2020robust}, and hence cannot leverage old patterns that may reappear.

The results of our experiments clearly demonstrate that our DA-LSTM model significantly outperforms commonly used baseline methods in terms of prediction accuracy, while also achieving a reasonable trade-off between performance and computational cost. In terms of prediction performance, the passive LSTM approach showed superior results compared to other methods in terms of MAPE and RMSE metrics, with notable average improvements across all households. The active approach also exhibited higher prediction performance, with average improvements for a significance level of $\tau=0.07$ and even higher improvements for $\tau=0.15$. Furthermore, our DA-LSTM method surpassed other commonly used baseline methods from the literature, such as RNN, ARIMA, and Bagging Regression, by a significant margin in terms of MAPE and RMSE for all households.

By comparing the performance-cost trade-off results of the different methods, we could gain deeper insights into which method might be best suited for different use cases. Specifically, we found that the active approach yielded a better overall trade-off performance than the passive approach. This suggests that the active approach could be more suitable in scenarios where computational cost is a determinative factor. Another important advantage of the active LSTM approach is its interpretability. Active detection provides additional information about change points in the time-series data, allowing for a better understanding of the underlying dynamics driving the consumption patterns. This interpretability can be particularly useful for power utility companies in real-world applications, where it is important to have insights into why and when changes in consumption occur. In contrast, passive adaptation provides more accurate predictive capabilities but at the cost of higher computational burden and limited interpretability due to the lack of information on changes in the data.

\subsection{Limitations and Outlook}
\label{limitations}

To assess the proposed framework, we evaluated the framework for the load forecasting use case with LSTM models. However, the framework is not limited to these settings, and the drift-adaptation methodology is model-agnostic and can be integrated with any ML-based predictor. Furthermore, the framework can be applied to other data types. In the multivariate case, the divergence metric will be calculated based on a multivariate probability density function. Additionally, the predictive model should support multiple inputs in the input layer that we would investigate in future work. For drift magnitude sensitivity, the TS metric could be a useful indicator that helps in selecting the most appropriate approach for the corresponding problem. However, some limitations may also affect the adoption of the appropriate approach, such as a threshold for minimum desired performance or budget allocations for costs.


\section{Conclusion}
\label{sec:conclusion}
\ac{dl} techniques have been exploited in the problem of interval load forecasting of residential households. Most existing solutions use offline learning, where the solution is built using historical data and deployed once it achieves good results with training data. This solution does not guarantee good performance after deployment since changes could occur and the solution would be obsolete. This paper proposes a drift-adaptive framework for LSTM networks (DA-LSTM) that can dynamically detect and adapt to changes. The main characteristic of the proposed framework is that it does not require fixing a drift threshold, since it evolves with time dynamically using the drift magnitude distribution. We integrate several detection strategies and apply them to real-world datasets. The evaluation is carried out in terms of prediction performance using MAPE and \ac{rmse} metrics, and the associated computational costs of using \ac{cpu} and \ac{gpu} resources. Additionally, the costs of using a cloud-based service (AWS) are calculated to quantify the deployment costs. The evaluation results demonstrate the efficiency of our solution compared to conventional LSTM and other popular baseline methods in terms of prediction performance. We also present an analysis of the trade-off between the performance and costs of each approach that would provide suggestions for adopting the appropriate approach in real-life problems. 

\section*{Acknowledgement}
This work has been partially funded by the Knowledge Foundation of Sweden (KKS) through the Synergy Project AIDA - A Holistic AI-driven Networking and Processing Framework for Industrial IoT (Rek:20200067). Additional funding has been provided by the Swedish Energy Agency through the AI4-ENERGI project (grant number 50246-1) and the project Solar Electricity Research Centre, Sweden (SOLVE), grant number 52693-1.

\balance
\bibliographystyle{unsrturl}
\bibliography{refs.bib}

\end{document}